\documentclass[10pt,twocolumn,letterpaper]{article}
\usepackage{multirow}
\usepackage{makecell}
\usepackage{subcaption}
\usepackage{amsmath}
\DeclareMathOperator*{\argmax}{arg\,max}

\usepackage{tabularx}
\usepackage{float}
\usepackage{pifont}
\usepackage{hhline}
\usepackage[table]{xcolor}
\usepackage{nicematrix}
\usepackage{balance}
\usepackage[pagenumbers]{cvpr} 

\definecolor{cvprblue}{rgb}{0.21,0.49,0.74}
\usepackage[pagebackref,breaklinks,colorlinks,allcolors=cvprblue]{hyperref}

\def\paperID{4120} 
\def\confName{CVPR}
\def\confYear{2025}

\title{CADRef: Robust Out-of-Distribution Detection via Class-Aware Decoupled Relative Feature Leveraging}

\author{Zhiwei Ling \and Yachen Chang \and Hailiang Zhao\thanks{Corresponding authors.} \and Xinkui Zhao \and Kingsum Chow$^*$ \and Shuiguang Deng\\
Zhejiang University\\
{\normalsize \url{{zwling, yachenchang, hliangzhao, zhaoxinkui, kingsum.chow, dengsg}@zju.edu.cn}}\\
}

\begin{document}
\maketitle
\begin{abstract}

\indent Deep neural networks (DNNs) have been widely criticized for their overconfidence when dealing with out-of-distribution (OOD) samples, highlighting the critical need for effective OOD detection to ensure the safe deployment of DNNs in real-world settings.
Existing post-hoc OOD detection methods primarily enhance the discriminative power of logit-based approaches by reshaping sample features, yet they often neglect critical information inherent in the features themselves.
In this paper, we propose the \underline{C}lass-\underline{A}ware \underline{Re}lative \underline{F}eature-based method (CARef), which utilizes the error between a sample’s feature and its class-aware average feature as a discriminative criterion.
To further refine this approach, we introduce the \underline{C}lass-\underline{A}ware \underline{D}ecoupled \underline{Re}lative \underline{F}eature-based method (CADRef), which decouples sample features based on the alignment of signs between the relative feature and corresponding model weights, enhancing the discriminative capabilities of CARef.
Extensive experimental results across multiple datasets and models demonstrate that both proposed methods exhibit effectiveness and robustness in OOD detection compared to state-of-the-art methods.
Specifically, our two methods outperform the best baseline by 2.82\% and 3.27\% in AUROC, with improvements of 4.03\% and 6.32\% in FPR95, respectively.

\end{abstract}    
\section{Introduction}
\label{sec:intro}

The remarkable progress in DNNs over the past few years has expanded their application across various domains \cite{lecun2015deep, DBLP:journals/tgrs/LiSFCGB19, resnet}. However, this success brings an equally significant challenge in terms of model reliability and safety. When exposed to out-of-distribution (OOD) samples, DNNs deployed in real-world settings frequently produce confidently incorrect predictions, failing to recognize when samples fall outside their classification capabilities \cite{MSP, ODIN, DBLP:conf/cvpr/NguyenYC15}. This vulnerability introduces substantial risks in safety-critical fields, such as autonomous driving \cite{autonomous_driving} and medical diagnostics \cite{medical}, where incorrect predictions could lead to severe consequences. A trustworthy DNN model must not only achieve high accuracy on in-distribution (ID) samples but also effectively identify and reject OOD samples \cite{DBLP:conf/cvpr/XueHZXLT24, DBLP:conf/cvpr/YuanHDHY24}. Therefore, the development of robust OOD detection methods has become an urgent priority for ensuring the safe deployment of DNNs \cite{MOS,DBLP:conf/cvpr/BaiHCJHZ24}.

\begin{figure}[t]
    \centering
    \begin{subfigure}{0.235\textwidth}
        \centering
        \includegraphics[width=\textwidth]{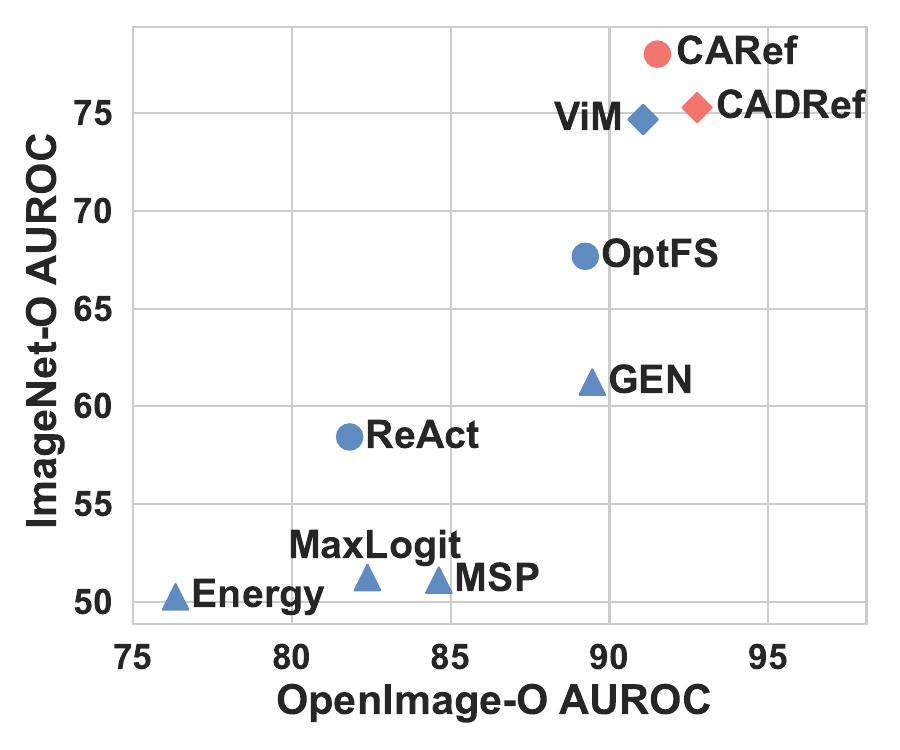}
        \caption{Performance of Methods}
        \label{fig:re_sub1}
    \end{subfigure}
    \hfill
    \begin{subfigure}{0.235\textwidth}
        \centering
        \includegraphics[width=\textwidth]{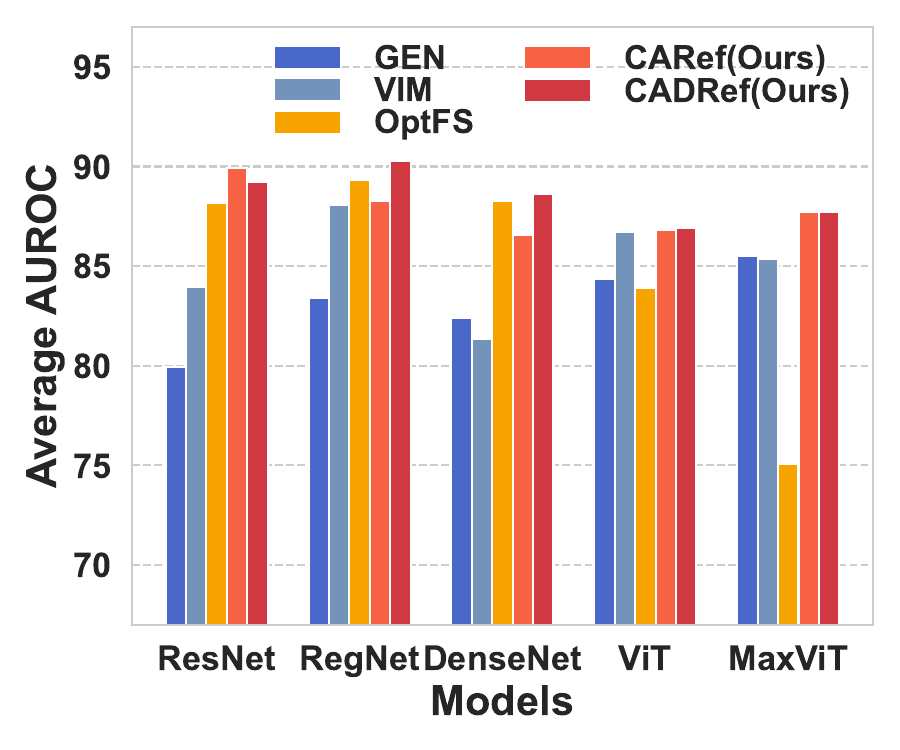}
        \caption{Performance across Models}
        \label{fig:re_sub2}
    \end{subfigure}
    \vspace{-0.2in}
    \caption{
    Performance of post-hoc OOD detection methods: 
    (a) shows the average AUROC of various methods tested on the ImageNet-O and OpenImage-O datasets, where $\triangle$, $\bigcirc$, and $\Diamond$ represent logit-based methods, feature-based methods, and methods that fuse logits and features, respectively.
    (b) presents the average AUROC of our methods compared to three SOTA methods across different model architectures on the ImageNet-1k benchmark.
    }
    \label{average}
    \vspace{-0.2in}
\end{figure}

Currently, the prevailing approach to OOD detection involves designing a post-hoc score function that assigns confidence scores to samples, where ID samples receive high scores, and OOD samples receive low scores to enable clear differentiation \cite{OptFS, ReAct, Energy, cvpr_test_time}. Literature on post-hoc OOD detection can be categorized into two main types: \textbf{Logit-based} methods and \textbf{Feature-based} methods \cite{survey1, survey2,DBLP:conf/nips/BehpourDLHGR23, DBLP:conf/nips/ChenLQWWX23}. As illustrated in Figure \ref{fig:re_sub1}, feature-based methods generally outperform logit-based methods to a certain extent, indicating a performance gap. However, Figure \ref{fig:re_sub2} shows that logit-based methods can achieve superior results on a few specific models. Despite the effectiveness of feature-based methods, they primarily focus on reshaping features and subsequently adjusting logits, often neglecting the rich information embedded within the features. This raises an important question: \textit{Can OOD detection methods be designed to integrate both features and logits while fully leveraging the information inherent in sample features?}

In this paper, we introduce a novel OOD detection algorithm called Class-Aware Relative Feature Leveraging (\textit{CARef}). Unlike other feature-based OOD detection methods, our approach specifically focuses on the rich information embedded within sample features. \textit{CARef} first aggregates the sample features in the ID training dataset by class to obtain the average features for each class. For a given test sample, we then compute the relative error between its features and the class-specific average features to determine whether it is an OOD sample. To establish a connection with logits, we further extend \textit{CARef} to the Class-Aware Decoupled Relative Feature (\textit{CADRef}) method, which decouples the sample’s relative features into two components. This decoupling is based on the sign alignment between the relative features and corresponding model weights according to their contribution to the maximum logit. \textit{CADRef} also incorporates advanced logit-based scoring methods to scale the relative errors of these two components, effectively amplifying the distinction between ID and OOD samples. In summary, this paper makes the following three key contributions:
\begin{itemize}[leftmargin=2em]
    \item 
    From the perspective of class-specific feature discrepancy, we propose a simple yet effective OOD detection method, \textit{CARef}, which calculates the relative error of a sample’s features in relation to class-aware features to determine the OOD score.
    \item 
    By leveraging the association between logits and features, we extend \textit{CARef} to \textit{CADRef}, decoupling sample features based on the alignment of signs between relative features and corresponding model weights. This extension also integrates logit-based OOD detection methods to scale positive errors.
    \item 
    Comprehensive experimental results across multiple datasets and architectures demonstrate that our methods achieve notable improvements while consistently exhibiting robustness across different architectures.
\end{itemize}

The rest of this paper is organized as follows: 
Section \ref{sec:preliminaries} provides a brief overview of the preliminaries in OOD detection.
Section \ref{sec:related_work} reviews the related OOD detection methods.
Section \ref{sec:approach} describes the design of our proposed methods (\textit{CARef} \& \textit{CADRef}), and Section \ref{sec:experiment} presents the experimental results. 
Finally, Section \ref{sec:conclusion} concludes the paper.
\section{Preliminaries}
\label{sec:preliminaries}
Consider a sample classification task with $c$ classes.
Given a DNN model $\mathcal{M}: \mathcal{X} \to \mathbb{R}^{c}$ trained on an ID training dataset $\mathcal{D}_\text{train}$, the prediction label for any sample $x \in \mathcal{D}_\text{test}$ is given by $y = \argmax_i \mathcal{M}(x)_i$.
We further define $\mathcal{F}: \mathcal{X} \to \mathbb{R}^{d}$ as the feature extractor, where $\mathcal{F}(x)$ represents the feature vector from the penultimate layer for the input $x$. 
Additionally, let $\mathcal{W}: \mathbb{R}^{d} \to \mathbb{R}^{c}$ and $\mathcal{B}: \mathbb{R}^{c} \to \mathbb{R}^{c}$ denote the weights and biases of the classifier, respectively. 
The logits $\mathcal{L}$ produced by the model for sample $x$ are computed as follows:
\begin{align}
\mathcal{L} &= \mathcal{M}(x) = \mathcal{W} \cdot \mathcal{F}(x) + \mathcal{B}, \\
\mathcal{T} &= \arg \max_i \mathcal{M}(x)_i.
\end{align}

Let $\mathcal{D}_{\text{I}}$ denote the ID dataset, while $\mathcal{D}_\text{O}$ stands for the OOD dataset.
The goal of OOD detection is to determine whether a given sample originates from the in-distribution dataset  $\mathcal{D}_{\text{I}}$ or the out-of-distribution dataset $\mathcal{D}_\text{O}$, effectively framing it as a binary classification task. This task is based on a scoring function $\textsc{Score}(\cdot;\cdot)$:
\begin{equation}
\left\{
\begin{array}{ll}
    x \sim \mathcal{D}_\text{O}, & \text{if } \textsc{Score}(\mathcal{M}; x) \leq \gamma, \\
    x \sim \mathcal{D}_{\text{I}}, & \text{if } \textsc{Score}(\mathcal{M}; x) > \gamma,
\end{array}
\right.
\label{ood_objective}
\end{equation}
where $\gamma$ is the threshold.
According to \eqref{ood_objective}, a sample is classified as an OOD sample if its score falls below $\gamma$; otherwise, it is classified as an ID sample. In real-world applications, once a sample is identified as an OOD sample, the DNN should abstain from making any predictions for it.
\section{Related Work}
\label{sec:related_work}

The design of post-hoc OOD score can be categorized into Logits-based ($\mathcal{S}_\text{logit}$) and Features-based ($\mathcal{S}_\text{feature}$) methods.

\subsection{Logit-based OOD-Detection Method}

\renewcommand{\arraystretch}{1.2}
\begin{table}[htbp]
\centering
\begin{tabular}{l|l}
\Xhline{1px}
\bf{Method} & \textbf{Score Equation} \\
\hline
\textit{MSP} \cite{MSP} & $\max(\mathrm{SoftMax}(\mathcal{L}))$\\
\textit{MaxLogit} \cite{MaxLogit}& $ \max(\mathcal{L})$\\
\textit{ODIN} \cite{ODIN}& $\max(\mathrm{SoftMax}(\tilde{\mathcal{L}}))$ \\
\textit{Energy} \cite{Energy}& $ T \cdot \log\sum(\exp(\mathcal{L} / T))$ \\
\textit{GEN} \cite{GEN}& $- \sum_{i=1}^M p_i^{\gamma}(1 - p_i)^{\gamma}$ \\
\Xhline{1px}
\end{tabular}
\caption{Score equations of logit-based OOD-detection methods}
\label{logit-based}
\end{table}

\textbf{Logit-based} methods analyze the logits produced by the model for each sample and design confidence scores such that in-distribution (ID) samples yield higher scores while OOD samples yield lower scores \cite{MSP,DBLP:conf/cvpr/ZhangX23, icml_logit_norm}. For example, Hendrycks \textit{et al.} propose the classic baseline method, \textit{MSP} \cite{MSP}, which uses the maximum value of the logits after applying the softmax function. Additionally, \textit{ODIN} \cite{ODIN} enhances detection by using temperature scaling and adding small perturbations to the input. Both \textit{Energy} \cite{Energy} and \textit{GEN} \cite{GEN} leverage the energy function and generalized entropy, respectively, to construct OOD scores, achieving notable improvements. For large-scale anomaly segmentation tasks, Hendrycks \textit{et al.} observed that using maximum logits can yield superior performance \cite{MaxLogit}. Table \ref{logit-based} summarizes the scoring equations for these methods.

\subsection{Feature-based OOD-Detection Method}

\renewcommand{\arraystretch}{1.2}
\begin{table}[htbp]
\centering
\begin{tabular}{l|l}
\Xhline{1px}
\bf{Method} & \textbf{Score Equation} \\
\hline
\textit{ReAct} \cite{ReAct} & $\mathcal{S}_\text{logit}(\mathcal{W} \cdot \min(\mathcal{F}(x), \tau) + \mathcal{B})$\\
\textit{DICE} \cite{DICE}& $\mathcal{S}_\text{logit}((\mathcal{W} \odot \mathcal{M}) \cdot \mathcal{F}(x) + \mathcal{B})$\\
\textit{ViM} \cite{VIM}& $-\alpha\| \mathcal{F}(x)^{P^{\perp}} \|_2$ +  ${\mathcal{S}_\text{logit}(\mathcal{L})}$  \\ 
\textit{ASH-S} \cite{ASH}& $\mathcal{S}_\text{logit}(\mathcal{W} \cdot (a_s \odot \mathcal{F}(x)) + \mathcal{B})$ \\
\textit{ASH-P} \cite{ASH}& $\mathcal{S}_\text{logit}(\mathcal{W} \cdot (a_p \odot \mathcal{F}(x)) + \mathcal{B})$ \\
\textit{ASH-B} \cite{ASH}& $\mathcal{S}_\text{logit}(\mathcal{W} \cdot (a_b \odot \mathcal{F}(x)) + \mathcal{B})$ \\
\textit{OptFS} \cite{OptFS}& $\mathcal{S}_\text{logit}(\mathcal{W} \cdot (\theta \odot \mathcal{F}(x)) + \mathcal{B})$\\
\Xhline{1px}
\end{tabular}
\caption{Score equations of feature-based OOD-detection methods}
\label{feature-based}
\end{table}

\textbf{Feature-based} methods operate on the feature layer, often by computing or modifying elements in the feature space \cite{LINE,ReAct,maha,DOS}. 
These approaches are founded on observing and exploiting the comparative stability of feature behaviors in ID samples relative to OOD samples, thereby enhancing the discriminative gap between ID and OOD samples. For example, 
\textit{ReAct} \cite{ReAct} removes outliers by truncating features that exceed a certain threshold $\tau$. Djurisic \textit{et al.} \cite{ASH} introduce a straightforward method that removes most of a sample’s features while applying adjustments to the remaining ones. Their method includes three variants, \textit{ASH-S}, \textit{ASH-P}, and \textit{ASH-B}, which apply different masks to features. These masks are defined as follows:
\begin{equation}
\vspace{-0.01in}
a_s, a_p, a_b = \left\{
\begin{array}{ll} 
\exp(\frac{s_o}{s_p}), 1, \frac{s_o}{n \cdot \mathcal{F}(x)_{i}},  & \text{if } \mathcal{F}(x)_{i} \geq \tau,\\
0, & \text{if } \mathcal{F}(x)_{i} < \tau,
\end{array}
\right.
\vspace{-0.01in}
\label{ash_equation}
\end{equation}
where $s_o$ and $s_p$ denote the sum of features before and after pruning, and $n$ represents the number of features retained.
\textit{Scale} \cite{ISH} makes a slight modification to \textit{ASH-S} by retaining pruned features. Moving away from heuristic masks, Zhao \textit{et al.} reformulate feature-shaping masks as an optimization problem \cite{OptFS}, where the objective is to maximize the highest logit values for training samples.
\textit{DICE} \cite{DICE} ranks weights according to their contributions on the ID training dataset, pruning those with lower values. In contrast, \textit{ViM} \cite{VIM} captures the deviation of features from the principal subspace $P$, making full use of the information embedded in the features. \textit{ViM} \cite{VIM} also includes the logits associated with ID classes, addressing part of the issues we identified. Table \ref{feature-based} summarizes these feature-based methods.

\section{Approach}
\label{sec:approach}

In this section, we first introduce \textit{CARef} in Subsection \ref{CARef}, which is designed to compute class-aware relative feature errors. Following this, we extend \textit{CARef} to \textit{CADRef} by introducing two essential modules: \textit{Feature Decoupling} and \textit{Error Scaling}.

To achieve a fine-grained decoupling, we separate a sample’s features into positive and negative components based on their contribution to the maximum logit. By analyzing the two resulting error components, \textit{CADRef} effectively mitigates the positive errors of samples with high $\mathcal{S}_\text{logit}$ values. Details on these two modules are provided in Subsections \ref{Feature Decoupling} and \ref{Error Scaling}, respectively.

\begin{figure}[htbp]
    \centering    \includegraphics[width=0.24\textwidth]{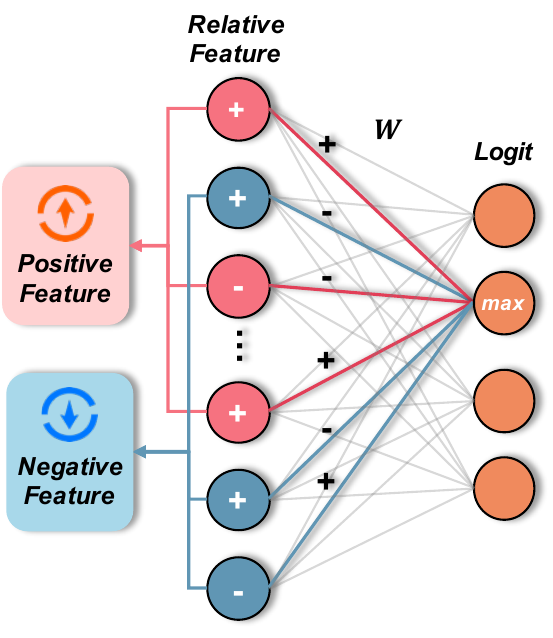}
    \caption{Example diagram of \textit{Feature Decoupling} operation of \textit{CADRef}. 
    Relative features refer to the gap between sample features and class-aware average features.
    The symbols $\textbf{+}$ and $\textbf{-}$ denote the sign of the corresponding values.}
    \label{fig:A1}
    \vspace{-0.1in}
\end{figure}

\subsection{Class-Aware Relative Feature Error}
\label{CARef}
We begin by extracting the features of the training samples, similar to other feature-based methods \cite{ReAct, DICE, OptFS}.
The key difference in our approach is that we group these features based on the predicted labels of the samples and compute the average feature vector for each class. For each $k \in \{ 1, 2, ..., c \}$, we define 
\begin{equation}
\vspace{-0.05in}
\mathcal{\overline{F}}^k = \frac{1}{n_k}\cdot \sum_{x \in \mathcal{D}_\text{train}} \mathbf{1} \Big(\mathcal{T}(x) =k \Big) \cdot \mathcal{F}(x),
\end{equation}
where $n_k$ is the number of samples with label $k$ in $\mathcal{D}_\text{train}$ and $\mathbf{1}(\cdot)$ is the indicator function.

To measure the deviation of individual sample features from their class averages, we propose calculating the relative error between a sample’s feature vector and its corresponding class centroid. Specifically, we compute the normalized  $l_1$-distance between the sample feature and the average feature, with normalization performed by the $l_1$-norm of the sample feature. The error formula and score function for \textit{CARef} are defined as:
\begin{equation}
\mathcal{E}(x) = \frac{\|\mathcal{F}(x) - \mathcal{\overline{F}}^{\mathcal{T}(x)}\|_{1}}{\|\mathcal{F}(x)\|_{1}}, \quad \textsc{Score}_{\textit{CARef}} \!=\! -\mathcal{E}(x).
\end{equation}
Our experimental results, presented in Table \ref{result_big_scale} and Table \ref{result_small_scale}, demonstrate that using class-aware relative feature error as a score function yields remarkable effectiveness.

\begin{figure*}[ht]
    \centering
    \begin{subfigure}{0.195\textwidth}
        \centering
        \includegraphics[width=\linewidth]{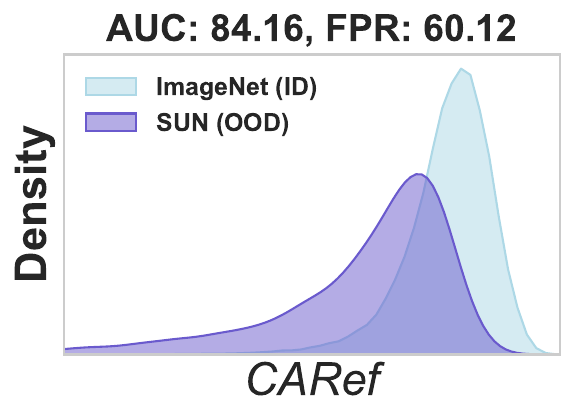}
        \caption{\textit{CARef}}
        \label{fig:caref}
    \end{subfigure}
    \hfill
    \begin{subfigure}{0.195\textwidth}
        \centering
        \includegraphics[width=\linewidth]{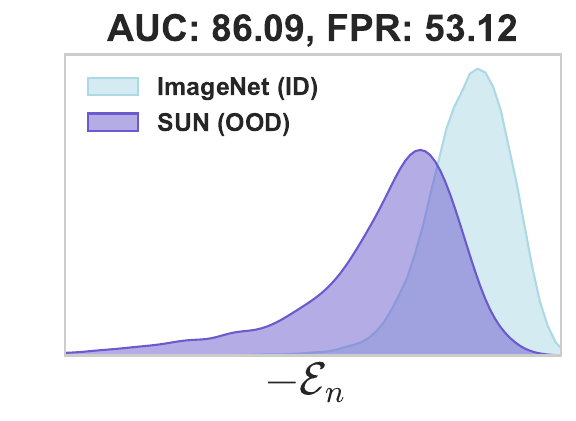}
        \caption{Negative Feature}
        \label{fig:sun_neg}
    \end{subfigure}
    \hfill
    \begin{subfigure}{0.195\textwidth}
        \centering
        \includegraphics[width=\linewidth]{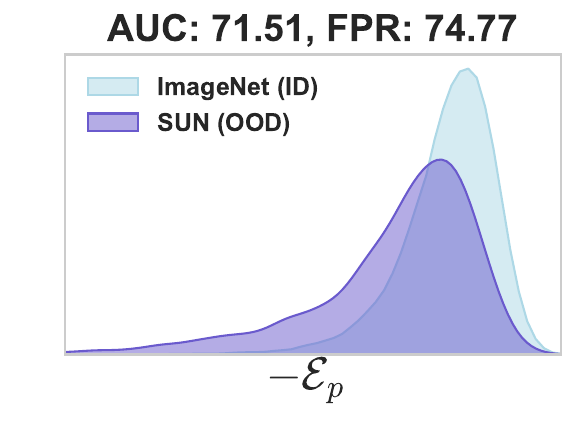}
        \caption{Positive Feature}
        \label{fig:sun_pos}
    \end{subfigure}
     \hfill
    \begin{subfigure}{0.195\textwidth}
        \centering
        \includegraphics[width=\linewidth]{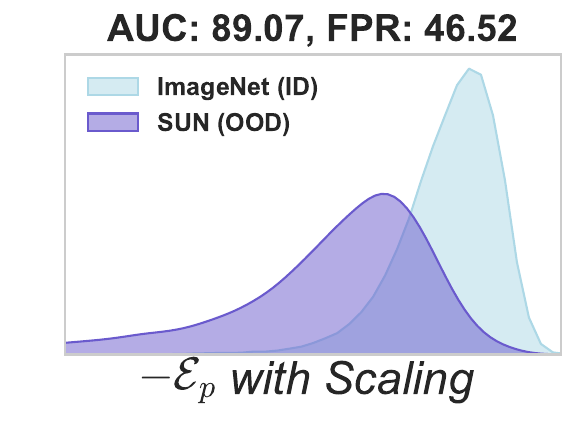}
        \caption{Scaled Positive Feature}
        \label{fig:sun_pos_wo}
    \end{subfigure}
     \hfill
    \begin{subfigure}{0.195\textwidth}
        \centering
        \includegraphics[width=\linewidth]{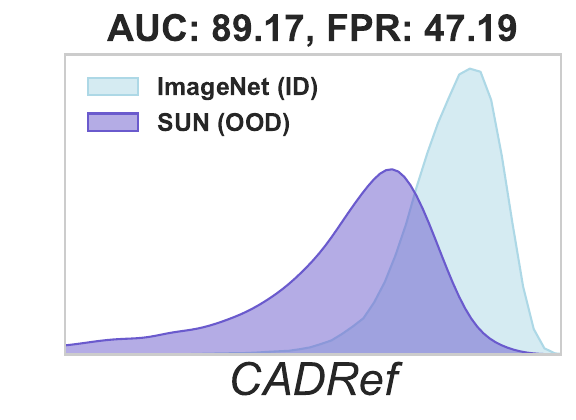}
        \caption{\textit{CADRef}}
        \label{fig:cadref}
    \end{subfigure}
    
    \caption{Detection results and score distributions on ImageNet-1k (blue) \cite{imagenet} and SUN (purple) \cite{sun} using DenseNet-201 \cite{resnet}.}
    \label{fig:sun}
\end{figure*}

\subsection{Feature Decoupling}
\label{Feature Decoupling}
As noted in \cite{VIM}, relying solely on sample features limits the effectiveness of OOD detection methods. To address this, we explore the relationship between features and logits in depth and conduct a fine-grained analysis of the features, which distinguishes \textit{CADRef} from \textit{CARef}. Based on an empirical observation, increasing the \textit{maximum} logit (i.e., $\max(\mathcal{L})$) of a sample tends to improve the performance of most logits-based methods. Therefore, we focus on the change in the maximum logit of the sample features relative to the class-aware average features.

Let $\mathcal{W}^\text{max}$ denote the weights corresponding to the maximum logit. As illustrated in Figure \ref{fig:A1}, the contribution to logit values depends on the alignment of signs between the weights and relative features. Specifically, a positive contribution to the maximum logit occurs only when the signs of weights and relative features align, while a mismatch in signs results in an antagonistic effect that diminishes the logit value. By influencing the maximum logit, these positive features also affect logit-based detection methods. We will analyze this relationship further in the next subsection.

Furthermore, the features of each sample can be divided into two parts: $\textsc{Pos} = \{i \mid \mathcal{W}^{max}_i \cdot \mathcal{F}(x)_i > 0\}$ and $\textsc{Neg} = \{i \mid \mathcal{W}^{max}_i \cdot \mathcal{F}(x)_i < 0\}$. 
The corresponding errors for these two parts are as follows:
\begin{align}
\mathcal{E}_p(x) &= \frac{\|\sum_{i \in \textsc{Pos}} (\mathcal{F}(x)_i - \mathcal{\overline{F}}^{\mathcal{T}(x)}_i)\|_{1}}{\|\mathcal{F}(x)\|_{1}}, 
\\
\mathcal{E}_n(x) &= \frac{\|\sum_{i \in \textsc{Neg}}(\mathcal{F}(x)_i - \mathcal{\overline{F}}^{\mathcal{T}(x)}_i))\|_{1}}{\|\mathcal{F}(x)\|_{1}}.
\end{align}

\subsection{Error Scaling}
\label{Error Scaling}
Decomposing  $\mathcal{E}(x)$ into $\mathcal{E}_p(x)$ and $\mathcal{E}_n(x)$ does not impact the relative error between samples. To investigate their individual contributions, we evaluated these components separately as OOD detection scores. Empirical results show that, compared to \textit{CARef}, using the positive error as the score substantially reduces OOD detection performance, while using the negative error as the score achieves nearly comparable performance. As shown in Figure \ref{fig:sun}, the AUROC and FPR95 of $\mathcal{E}_p$ are approximately 13\% and 15\% lower than those of \textit{CARef}, respectively. 
In contrast, $\mathcal{E}_n$ shows a 2.03\% and 7.00\% increase. 
Due to the poor classification performance of the positive error, we conclude that it plays a harmful role in \textit{CARef}’s coupling form, highlighting the need to focus on enhancing $\mathcal{E}_p$.

\begin{figure}[htbp]
    \centering
    \begin{subfigure}{0.235\textwidth}
        \centering
        \includegraphics[width=\linewidth]{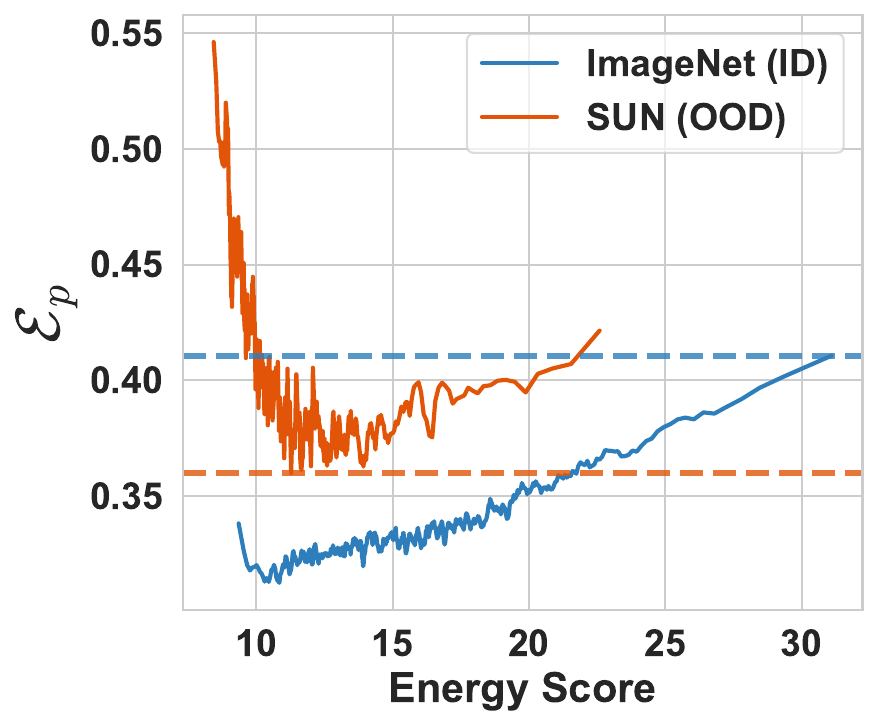}
        \caption{Positive Feature}
        \label{fig:energy_p}
    \end{subfigure}
    \hfill
    \begin{subfigure}{0.235\textwidth}
        \centering
        \includegraphics[width=\linewidth]{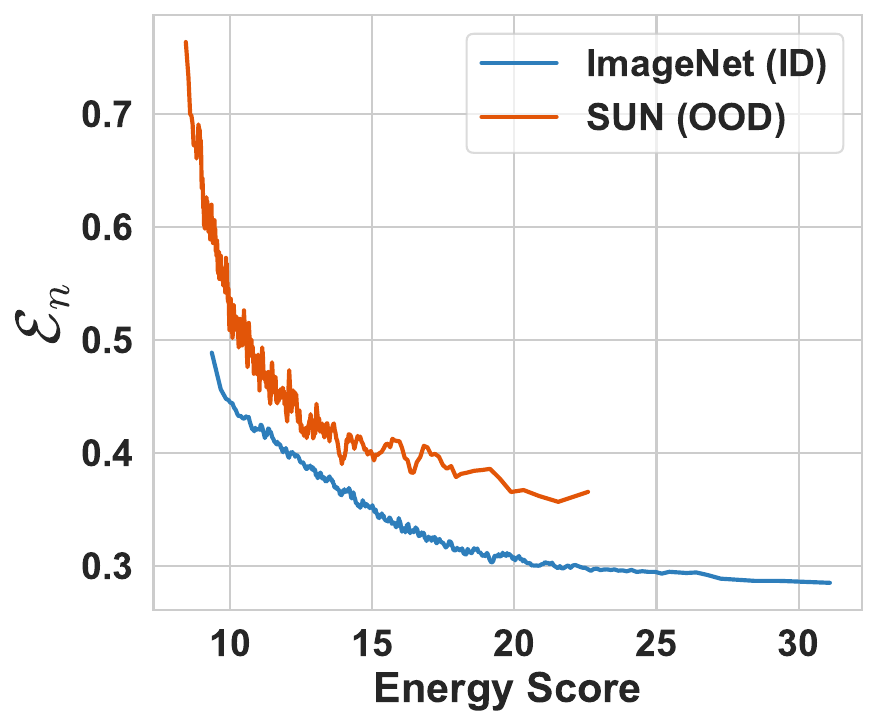}
        \caption{Negative Feature}
        \label{fig:energy_n}
    \end{subfigure}
    \\
    \begin{subfigure}{0.235\textwidth}
        \centering
        \includegraphics[width=\linewidth]{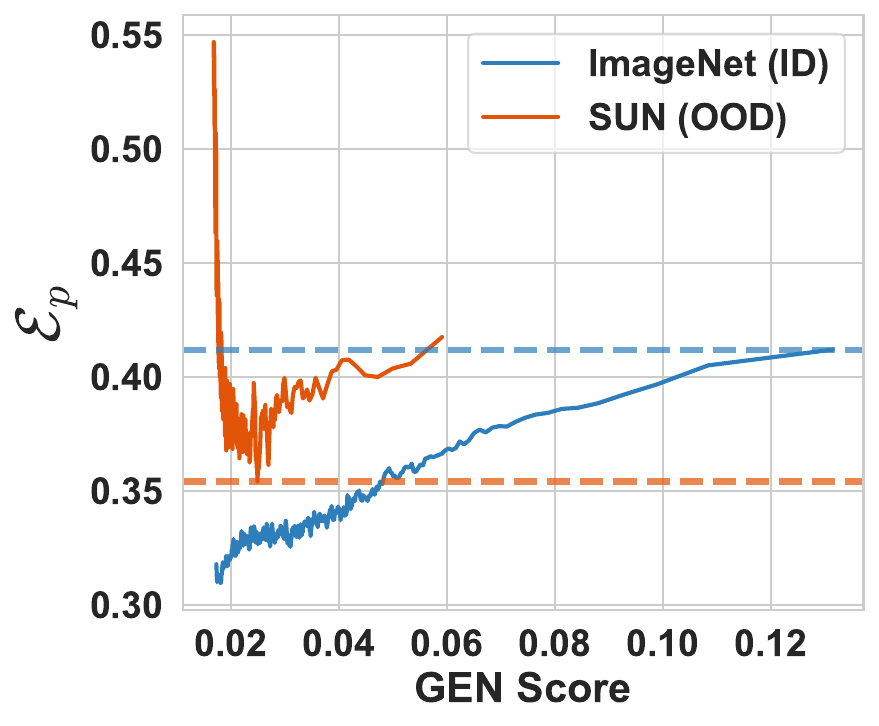}
        \caption{Positive Feature}
        \label{fig:gen_p}
    \end{subfigure}
    \hfill
    \begin{subfigure}{0.235\textwidth}
        \centering
        \includegraphics[width=\linewidth]{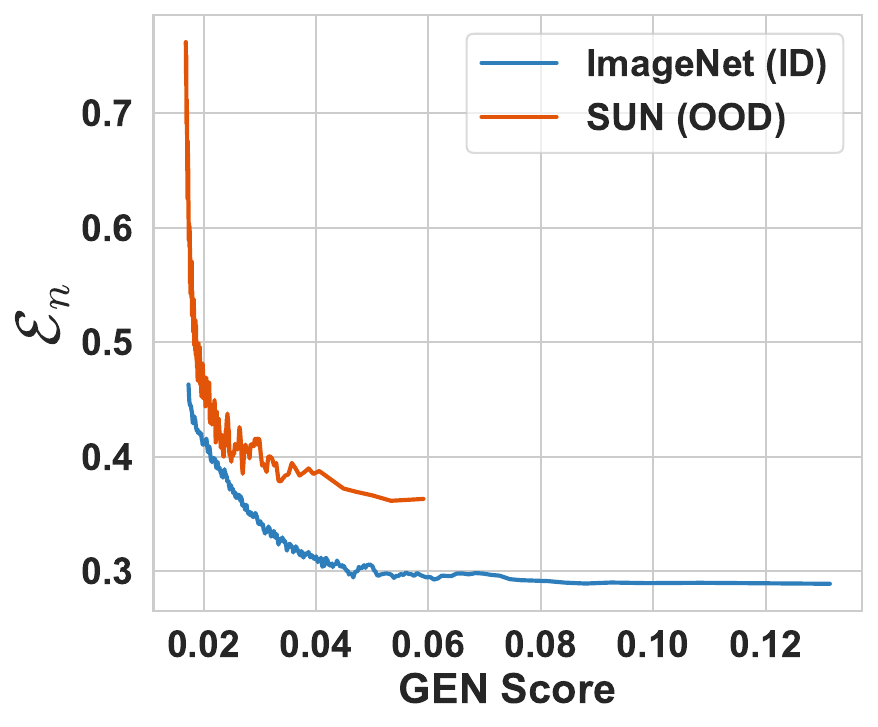}
        \caption{Negative Feature}
        \label{fig:gen_n}
    \end{subfigure}
    \caption{Score and error distribution of ID/OOD samples.}
\end{figure}

\renewcommand{\arraystretch}{1.4}
\begin{table*}[ht]
\centering
\setlength{\tabcolsep}{0.7mm}
\begin{NiceTabular}{c|cc|cc|cc|cc|cc|cc|cc|cc }
\CodeBefore
\rectanglecolor[gray]{0.9}{1-16}{14-17}
\Body
\Xhline{1px}
\multirow{2}{*}{\bf{Method}} 
& \multicolumn{2}{c|}{\bf{ResNet}} & \multicolumn{2}{c|}{\bf{RegNet}} & \multicolumn{2}{c|}{\bf{DenseNet}} & \multicolumn{2}{c|}{\bf{ViT}} & \multicolumn{2}{c|}{\bf{Swin}} & \multicolumn{2}{c|}{\bf{ConvNeXt}} & \multicolumn{2}{c|}{\bf{MaxViT}} & \multicolumn{2}{c}{\cellcolor[gray]{0.9}\bf{Average}}  \\ 
\hhline{~*{14}{-}|*{2}{-}}
& AU$\uparrow$          & FP$\downarrow$         & AU$\uparrow$         & FP$\downarrow$   & AU$\uparrow$         & FP$\downarrow$      & AU$\uparrow$           & FP$\downarrow$ & AU$\uparrow$          & FP$\downarrow$ & AU$\uparrow$  & FP$\downarrow$ & AU$\uparrow$           & FP$\downarrow$& AU$\uparrow$           & FP$\downarrow$ \\ \hline 
\textit{MSP} \cite{MSP} &74.14 & 70.44 & 78.34 & 69.98 & 77.76 & 67.63 & 79.33 & 66.29 & 77.81 & 67.12 & 76.23 & 65.24 & 80.36 &63.50&77.71 & 67.17\\
\textit{MaxLogit} \cite{MaxLogit} & 79.64 & 65.00 & 81.71 & 63.75 & 81.16 & 61.56 & 75.68 & 65.40 & 70.85 & 67.68 & 68.76 & 72.66 & 77.18& 57.28& 76.43 &64.76\\
\textit{ODIN} \cite{ODIN}  & 79.28 & 60.32 & 80.57 & 58.66 & 78.90 & 60.75 & 61.45 & 94.33 & 55.22 & 91.02 & 51.09 & 89.33 & 64.60 & 81.46 &67.30 &76.55\\
\textit{Energy} \cite{Energy} & 79.85 & 64.53 & 81.55 & 64.11 & 81.09 & 61.25 & 71.04 & 70.23 & 63.18 & 75.96 & 52.91 & 91.09 & 73.01 &62.46 &71.80 &69.95 \\
\textit{GEN} \cite{GEN} & 79.95 & 66.70 & 83.41 & 63.46 & 82.39 & 63.01 & 84.37 & \underline{59.58} & 83.72 & \textbf{55.33} & 82.69 & \underline{54.90} & 85.50 &51.22& 83.15&59.17 \\
\textit{ReAct} \cite{ReAct} & 86.03 & 44.09 & 86.62 & 46.30 & 78.46 & 64.18 & 79.65 & 69.99 & 81.91 & 66.57 & 78.15 & 76.87 & 63.78 & 78.91 &79.23 & 63.84\\
\textit{DICE} \cite{DICE} & 82.48 & 47.90 & 77.92 & 68.89 & 79.20 & 59.04 & 71.65 & 88.51 & 45.79 & 86.76 & 45.12 & 89.77 & 63.20 &76.78 &66.48 & 73.95\\
\textit{ASH-S} \cite{ASH} & \textbf{90.11} & \textbf{35.52} & \underline{89.37} & \textbf{38.01} & 87.56 & \textbf{43.80} & 18.06 & 99.62 & 19.18 & 99.35 & 20.30 & 98.23 & 55.26 & 86.30&54.26 &71.55 \\
\textit{OptFS} \cite{OptFS} & 88.15 & 42.47 & 89.30 & 42.97 & \underline{88.25} & 47.88 & 83.89 & 65.99 & 84.71 & 65.03 & 85.06 & 61.17 &75.09 & 70.84&84.92 &56.62 \\
\textit{ViM} \cite{VIM} & 83.96 & 65.85 & 88.08 & 56.13 & 81.33 & 72.66 & 86.72 & \textbf{49.36} & 83.96 & 63.63 & 84.24 & 58.57 & 85.34 &53.56 &84.80 &59.97 \\
\hline
\textit{CARef} & \underline{89.94} & 40.91 & 88.27 & 50.68 & 86.55 & 52.59 & \underline{86.84} & 60.48 & \underline{86.92} & 58.65 & \textbf{87.95} & \textbf{54.09} & \underline{87.70} & \underline{50.75} & \underline{87.74} & \underline{52.59}\\
\textit{CADRef} & 89.24 & \underline{40.68} & \textbf{90.26} & \underline{42.34} & \textbf{88.64} & \underline{45.25} & \textbf{86.91} & 60.05 & \textbf{87.10} & \underline{57.38} & \underline{87.48} & 56.92 & \textbf{87.73} & \textbf{49.49} & \textbf{88.19} &\textbf{50.30} \\
\Xhline{1px}
\end{NiceTabular}
\caption{Results of OOD detection on ImageNet-1k benchmark. 
$\uparrow$ indicates that higher values are better, while $\downarrow$ indicates that lower values are better.
All values are percentages, with the best and second-best results being \textbf{highlighted} and \underline{underlined}, respectively.}
\label{result_big_scale}
\end{table*}

Since $\mathcal{E}_p$ is closely related to logit-based methods, we explore the relationship between $\mathcal{S}_\text{logit}$ and $\mathcal{E}_p$ for a sample, using the \textit{Energy} \cite{Energy} and \textit{GEN} \cite{GEN} scores as examples. Results for other methods are provided in the supplementary materials. As shown in Figure \ref{fig:energy_p} and Figure \ref{fig:gen_p}, we observe that $\mathcal{E}_p$ effectively distinguishes between ID and OOD samples at lower $\mathcal{S}_\text{logit}$ values. However, at higher $\mathcal{S}_\text{logit}$ values, $\mathcal{E}_p$ for ID and OOD samples shows substantial overlap, which reduces OOD detection performance. To address this issue, we propose \textit{using the ratio of $\mathcal{E}_p$ to $\mathcal{S}_\text{logit}$ instead of $\mathcal{E}_p$ alone}. When ID and OOD samples have similar $\mathcal{E}_p$ values, the higher $\mathcal{S}_\text{logit}$ of ID samples reduces this ratio, while for OOD samples, the effect is reversed. For the negative error, as shown in Figure \ref{fig:energy_n} and Figure \ref{fig:gen_n}, we observe the opposite phenomenon. There is no overlap in $\mathcal{E}_n$ between ID and OOD samples at high $\mathcal{S}_\text{logit}$, while some overlap exists at low $\mathcal{S}_\text{logit}$. This overlap does not impact the detection performance of the negative error, as low $\mathcal{S}_\text{logit}$ values indicate that distinguishing these samples is challenging for any detection method. Therefore, modifying the negative error is deemed unnecessary for performance enhancement. 

Comparing Figure \ref{fig:sun_pos} with Figure \ref{fig:sun_pos_wo}, we can find that scaling the positive error significantly enhances the separation between ID and OOD samples.
Finally, we consider a fusion form of the positive and negative errors.
To ensure consistency in the magnitude of both errors, we also apply a constant decay to the negative error, with the constant term set to the average score of all ID training samples:
\begin{equation}
\overline{\mathcal{S}}_\text{logit} = \frac{1}{n} \cdot \! \! \sum_{x \in \mathcal{D}_\text{train}} \mathcal{S}_\text{logit}(x).
\end{equation}
The final score formula of our \textit{CADRef} is as follows:
\begin{align}
\mathcal{E}(x) \! = \! \frac{\mathcal{E}_p(x)}{\mathcal{S}_\text{logit}(x)} \! + \! \frac{\mathcal{E}_n(x)}{\overline{\mathcal{S}}_\text{logit}}, \quad \textsc{Score}_{\textit{CADRef}} \!= \! - \mathcal{E}(x).
\end{align}

As shown in Figure \ref{fig:cadref}, the fusion of both errors further enhances the distinction between ID and OOD samples compared to using each error individually.

\section{Experiments}
\label{sec:experiment}
In this section, we conduct extensive experiments across multiple datasets and models to evaluate the performance of our two methods, and compare them with state-of-the-art OOD detection methods, all of which are implemented by PyTorch \cite{Pytorch}.
\subsection{Setup}

\noindent{\bf{Datasets}}.
We conduct experiments on both large-scale and small-scale benchmarks. For the large-scale benchmark, we use ImageNet-1k \cite{imagenet} as the in-distribution (ID) dataset and evaluate performance across six commonly used OOD datasets: iNaturalist \cite{inaturalist}, SUN \cite{sun}, Places \cite{places}, Texture \cite{texture}, OpenImage-O \cite{VIM}, and ImageNet-O \cite{imagenet-o}. For the small-scale benchmark, we use CIFAR-10 \cite{cifar} and CIFAR-100 \cite{cifar} as ID datasets, with performance evaluated on six widely used OOD datasets: SVHN \cite{svhn}, LSUN-Crop \cite{lsun}, LSUN-Resize \cite{lsun}, iSUN \cite{isun}, Texture \cite{texture}, and Places \cite{places}. All OOD datasets have been resized to match the dimensions of the ID datasets.

\noindent{\bf{Models architectures}}. 
To validate the robustness of the proposed methods, we conduct experiments using several well-known model architectures, including convolutional neural networks (CNNs) and vision transformers. For ImageNet-1k, we use four representative CNN-based models: ResNet-50 \cite{resnet}, RegNetX-8GF \cite{regnet}, DenseNet-201 \cite{densenet}, and ConvNeXt-B \cite{convnext}, as well as three transformer-based models: ViT-B/16 \cite{vit}, Swin-B \cite{swin}, and MaxViT-T \cite{maxvit} to ensure a comprehensive evaluation. The pre-trained model weights are sourced from the PyTorch model zoo \cite{Pytorch}. For experiments on CIFAR-10 and CIFAR-100, we use DenseNet-101 \cite{densenet} with pre-trained weights provided by prior work \cite{DICE}.

\renewcommand{\arraystretch}{1.3}
\begin{table*}[htbp]
\centering
\setlength{\tabcolsep}{1.2mm}
\begin{NiceTabular}{c c|cc|cc|cc|cc|cc|cc|cc}
\CodeBefore
\rectanglecolor[gray]{0.9}{1-15}{26-16}
\Body
\Xhline{1px}
\multirow{2}{*}{}
&\multirow{2}{*}{\bf{Methods}} 
& \multicolumn{2}{c|}{\bf{SVHN}} & \multicolumn{2}{c|}{\bf{LSUN-C}} & \multicolumn{2}{c|}{\bf{LSUN-R}} & \multicolumn{2}{c|}{\bf{iSUN}} & \multicolumn{2}{c|}{\bf{Textures}} & \multicolumn{2}{c|}{\bf{Places}} & \multicolumn{2}{c}{\bf{Average}}  \\ 
\hhline{~~*{12}{-}|*{2}{-}}

& & AU$\uparrow$ &FP$\downarrow$         & AU$\uparrow$   & FP$\downarrow$ & AU$\uparrow$ & FP$\downarrow$  & AU$\uparrow$   & FP$\downarrow$ & AU$\uparrow$  & FP$\downarrow$ & AU$\uparrow$  & FP$\downarrow$ & AU$\uparrow$  & FP$\downarrow$ \\ \hline 
\multirow {12}{*}{\rotatebox{90}{\textit{CIFAR-10}}}&\textit{MSP} \cite{MSP}& 93.56 & 47.19 & 93.42 & 47.09 & 94.54 & 42.07 & 94.49 & 42.53 & 88.24 & 63.88 & 90.02 & 60.01 & 92.38 & 50.46 \\
&\textit{MaxLogit} \cite{MaxLogit}& 94.32 & 37.79 & 97.22 & 16.31 & 98.12 & 9.41 & 98.05 & 10.08 & 86.65 & 56.57 & 93.61 & 34.82 & 94.66 & 27.50 \\
&\textit{ODIN} \cite{ODIN} & 92.88 & 39.95 &96.02 & 21.34& \textbf{99.29} & \textbf{3.09} &\textbf{99.19} &\textbf{3.79} & 86.16 & 53.22 &  92.57 & 38.81 & 94.35& 26.70 \\
&\textit{Energy} \cite{Energy}& 94.19 & 38.71 & 97.29 & 15.55 & 98.18 & 8.70 & 98.11 & 9.42 & 86.56 & 56.66 & 93.67 & 33.92 & 94.67 & 27.16 \\
&\textit{GEN} \cite{GEN}& 95.19 & 30.75 & 96.99 & 18.29 & 97.93 & 11.29 & 97.87 & 11.93 & 88.87 & 54.00 & 93.34 & 36.30 & 95.03 & 27.09 \\
&\textit{ReAct} \cite{ReAct}& 66.05 & 97.18 & 78.03 & 87.24 & 84.86 & 71.13 & 83.77 & 73.66 & 68.08 & 90.85 & 75.53 & 83.72 & 76.05 & 83.96 \\
&\textit{DICE} \cite{DICE}& 94.96 & 27.74 & \underline{98.31} & \underline{8.86} & 99.05 & 4.22 & 98.99 & 5.16 & 87.33 & 45.33 & \textbf{93.86} & \textbf{31.84} & 95.42 & 20.52 \\
&\textit{ASH-S} \cite{ASH}& 98.73 & 6.16 & 98.13 & 9.67 & 98.91 & 4.84 & 98.87 & 5.13 & 95.29 & 23.58 & 93.58 & \underline{32.32} & \underline{97.25} & \underline{13.62} \\
&\textit{OptFS} \cite{OptFS}& 96.01 & 24.35 & 96.99 & 18.09 & 98.11 & 9.31 & 98.00 & 10.25 & 94.29 & 32.96 & 92.33 & 40.15 & 95.95 & 22.52 \\
&\textit{ViM} \cite{VIM}& 98.45 & 8.65 & 97.36 & 15.39 & \underline{99.28} & \underline{3.17} & 99.12 & 4.50 & 96.18 & 20.33 & 89.80 & 54.48 & 96.70 & 17.75 \\
\hhline{~*{13}{-}|*{2}{-}}
&\textit{CARef} & \underline{99.11} & \underline{4.66} & 98.22 & 9.51 & 98.93 & 5.20 & 98.77 & 6.11 & \textbf{96.79} & \textbf{16.29} & 91.59 & 41.19 & 97.23 & 13.83 \\
&\textit{CADRef}& \textbf{99.17} & \textbf{4.16} & \textbf{98.66} & \textbf{7.02} & 99.23 & 3.61 & \underline{99.13} & \underline{4.34} & \underline{96.71} & \underline{17.52} & \underline{93.72} & 32.74 & \textbf{97.77} & \textbf{11.56} \\
\Xhline{1px}
\multirow {12}{*}{\rotatebox{90}{\textit{CIFAR-100}}}&\textit{MSP} \cite{MSP} & 75.19 & 82.02 & 78.63 & 76.44 & 67.13 & 87.28 & 68.49 & 88.00 & 71.20 & 85.19 & 70.84 & 85.28 & 71.91 & 84.04 \\
&\textit{MaxLogit} \cite{MaxLogit} & 81.42 & 86.17 & 87.90 & 58.91 & 77.41 & 76.05 & 76.54 & 79.13 & 71.14 & 84.45 & 76.18 & 79.82 & 78.43 & 77.42 \\
&\textit{ODIN} \cite{ODIN}& 80.33 & 86.53 & 89.10 & 51.85 & 86.41 &56.66 & 85.78 & 59.03 & 73.57 & 80.49 & 76.83 & 79.80 & 82.00 & 69.06 \\
&\textit{Energy} \cite{Energy}& 81.30 & 88.03 & 88.11 & 58.19 & 77.77 & 75.17 & 76.79 & 78.61 & 70.99 & 85.00 & 76.21 & 79.95 & 78.53 & 77.49 \\
&\textit{GEN} \cite{GEN}& 80.97 & 78.89 & 83.72 & 70.82 & 71.51 & 84.11 & 72.00 & 85.15 & 74.26 & 83.68 & 73.88 & 83.36 & 76.06 & 81.00 \\
&\textit{ReAct} \cite{ReAct}& 69.13 & 96.75 & 78.84 & 77.21 & 86.44 & 68.03 & 82.86 & 74.78 & 67.15 & 92.07 & 59.99 & 89.72 & 74.07 & 83.09 \\
&\textit{DICE} \cite{DICE}& 88.18 & 60.06 & 92.98 & 36.40 & 88.23 & 55.03 & 88.50 & 52.49 & 77.22 & 61.27 & \textbf{81.18} & \textbf{73.89} & 86.05 & 56.52 \\
&\textit{ASH-S} \cite{ASH}& 95.79 & 24.75 & \underline{94.14} & \underline{29.98} & 89.54 & 54.06 & 90.93 & 48.15 & 92.11 & 34.60 & \underline{79.22} & \underline{76.96} & 90.29 & 44.75 \\
&\textit{OptFS} \cite{OptFS}& 84.96 & 73.61 & 90.01 & 47.98 & 83.61 & 69.52 & 84.39 & 70.56 & 85.63 & 61.64 & 74.37 & 80.96 & 83.83 & 67.38 \\
&\textit{ViM} \cite{VIM}& 93.57 & 35.05 & 92.76 & 40.06 & \textbf{95.50} & \textbf{24.65} & \textbf{95.63} & \textbf{23.22} & \textbf{95.89} & \textbf{19.75} & 75.61 & 83.89 & \textbf{91.49} & \textbf{37.77} \\
\hhline{~*{13}{-}|*{2}{-}}
&\textit{CARef} & \textbf{96.83} & \textbf{17.41} & 90.96 & 40.74 & 89.25 & 52.69 & 91.08 & 45.32 & 93.99 & \underline{25.48} & 67.92 & 88.30 & 88.34 & 44.99 \\
&\textit{CADRef} & \underline{96.69} & \underline{18.28} & \textbf{94.70} & \textbf{27.22} & \underline{90.26} & \underline{47.45} & \underline{91.59} & \underline{42.10} & \underline{94.13} & 28.72 & 75.91 & 78.30 & \underline{90.55} & \underline{40.34} \\
\Xhline{1px}
\end{NiceTabular}
\caption{
Results of OOD detection on CIFAR benchmarks. 
$\uparrow$ indicates that higher values are better, while $\downarrow$ indicates that lower values are better.
All values are percentages, with the best and second-best results being \textbf{highlighted} and \underline{underlined}, respectively.}
\label{result_small_scale}
\end{table*}

\noindent{\bf{Baselines}}.
We evaluate our two proposed methods, with \textit{CADRef} leveraging the \textit{Energy} \cite{Energy} score as the source for \textit{Error Scaling}. 
For comparison, we implement ten baseline methods for OOD detection, covering both logit-based and feature-based approaches. 
The logit-based methods include \textit{MSP} \cite{MSP}, \textit{MaxLogit} \cite{MaxLogit}, \textit{ODIN} \cite{ODIN}, \textit{Energy} \cite{Energy} and \textit{GEN} \cite{GEN}.
Meanwhile, the feature-based methods comprise \textit{ReAct} \cite{ReAct},  \textit{DICE} \cite{DICE},  \textit{ASH-S} \cite{ASH}, \textit{OptFS} \cite{OptFS} and \textit{ViM} \cite{VIM}.
Note that all feature-based methods use \textit{energy} as the score function.
Details of the hyperparameters for each baseline are provided in the \textit{supplementary materials}.

\noindent{\bf{Evaluation metrics.}}
We evaluate the OOD detection performance using two standard metrics, consistent with prior works \cite{MSP, DBLP:conf/cvpr/TangHPCZWT24}: area under the receiver operating characteristic curve (AUROC) and false positive rate at a true positive rate of 95\% (FPR95). 
Higher AUROC values and lower FPR95 values indicate better OOD detection performance.

\subsection{Comparison with SOTA Methods}

\noindent{\bf{On ImageNet-1k Benchmark.}}
Table \ref{result_big_scale} presents the experimental results on ImageNet-1k.
We also provide detailed results for all datasets in the \textit{supplementary materials}.
From the average results among all models, both \textit{CARef} and \textit{CADRef} remain in the top two positions. 
Compared to the best baseline, \textit{CADRef} improves the AUROC by 3.27\% and reduces the FPR95 by 6.32\%, while \textit{CARef} also improves the AUROC by 2.82\% and reduces the FPR95 by 4.03\%.
We group the baselines for detailed analysis:

\begin{itemize} 
    \item \textbf{vs. feature shaping-based methods:} Experimental results demonstrate that feature shaping-based methods exhibit strong performance on specific architectures, notably ResNet-50, RegNetX-8GF, and DenseNet-201. 
    For example, \textit{ASH-S} achieves state-of-the-art performance on ResNet-50 and reports the lowest FPR95 values on both RegNetX-8GF and DenseNet-201. 
    However, for other model architectures, the performance of these methods significantly declines, with the \textit{ASH-S} method even collapsing (AUROC below 50\% on ViT-B/16, Swin-B, and ConvNeXt-B).
    This observation indicates that feature shaping-based methods exhibit architecture-specific behavior and lack cross-model robustness. 
    In contrast, our methods demonstrate outstanding performance on individual models, almost always ranking in the top two, further highlighting the robustness.
    To address the issue of performance collapse, \textit{OptFS} discards the heuristic mask design based on empirical methods and instead focuses on automating the optimization of masks across different models. 
    While \textit{OptFS} substantially improves cross-architecture robustness, as evidenced in Table \ref{result_big_scale}, its performance remains inferior to both \textit{CARef} and \textit{CADRef}.

    \item \textbf{vs. logit-based methods:} According to the Table \ref{result_big_scale}, compared to feature shaping-based methods, most logit-based methods do not perform well, except for \textit{GEN}. 
    However, these methods do not suffer from similar collapses on certain models as feature shaping-based methods, making them more generalizable compromise solutions.
    Additionally, logit-based methods do not require extra ID training data, which gives them an advantage in terms of computational resources.

    \item \textbf{vs. \textit{ViM}:} We focus especially on the comparison with \textit{ViM}, since it similarly utilizes feature information and logit-based scores, making it a method of the same category as \textit{CADRef}.
    As shown in the Table \ref{result_big_scale}, \textit{CADRef} outperforms \textit{ViM} in all other cases except for the FPR95 on ViT-B/16.
    On DenseNet, the fact that the effect of \textit{ViM} is closely approximated by \textit{Energy} (logit-based component) suggests that \textit{ViM} does not fully utilize feature information.
    Our method demonstrates that, compared to projecting into other spaces, ID and OOD samples exhibit significant separability within the feature space.
\end{itemize}

\noindent{\bf{On CIFAR Benchmarks.}}
Table \ref{result_small_scale} also shows the experimental results on CIFAR-10 and CIFAR-100 benchmarks.
Our proposed \textit{CADRef} demonstrates superior performance on CIFAR-10, achieving state-of-the-art results, while maintaining competitive performance on CIFAR-100 with only marginal differences from the best baseline method.
We also observe a decline in the performance of \textit{CARef} on the CIFAR benchmarks. 
This phenomenon can be attributed to the reduced feature dimensionality in small-scale datasets, which potentially compromises the precision of relative error calculations.

\subsection{Impact of various logit-based methods}
\begin{figure}[htbp]
    \centering
    \includegraphics[width=0.475\textwidth]{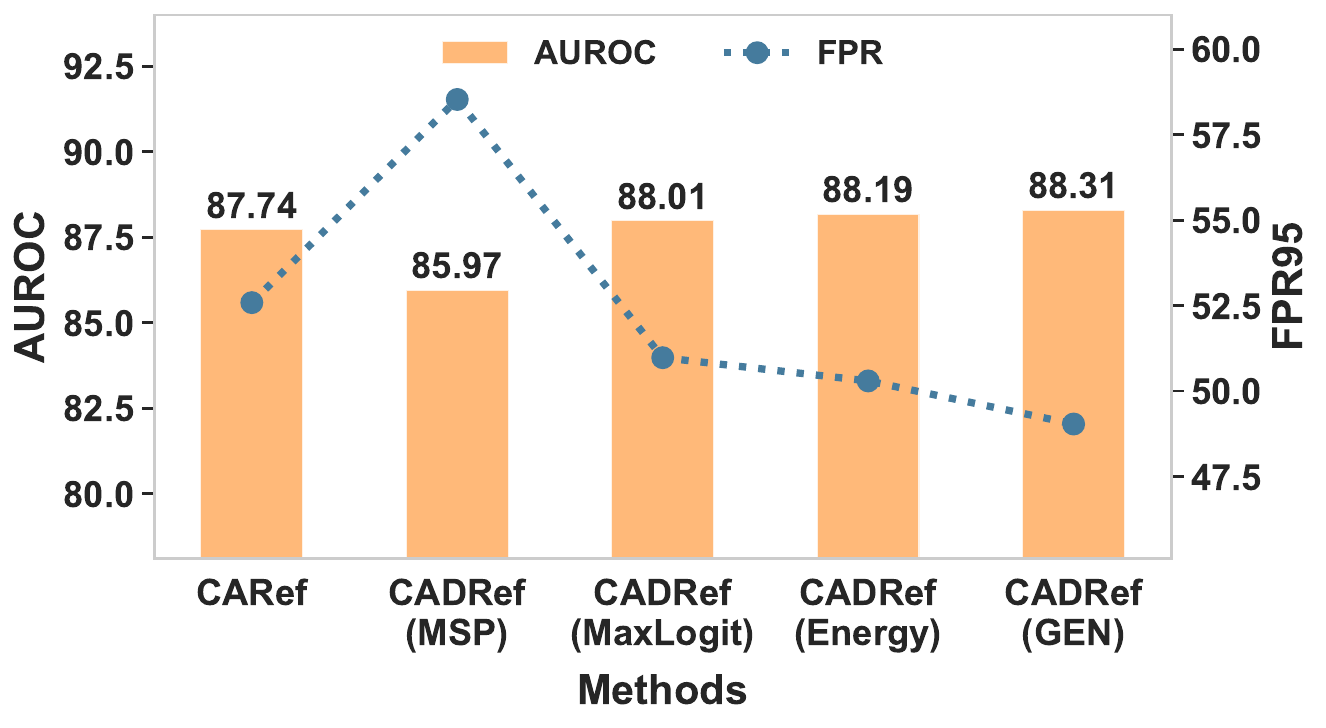}
    \caption{The Impact of various logit-based methods on \textit{CADRef}}
    \label{impact}
\end{figure}

Figure \ref{impact} provides the performance of \textit{CADRef} on the ImageNet-1k benchmark,
using \textit{MSP}, \textit{Maxlogit}, \textit{Energy}, and \textit{GEN} as the logit-based scores, respectively.
The results demonstrate that \textit{CADRef+GEN} achieves superior performance in both AUROC and FPR95 metrics among all logit-based variants, which aligns with the exceptional performance of \textit{GEN} previously observed in Table \ref{result_big_scale}.
Furthermore, the performance of both \textit{CADRef+MaxLogit} and \textit{CADRef+Energy} surpasses all baselines in Table \ref{result_big_scale}.
We also observe an interesting phenomenon in Figure \ref{impact} that deserves exploration in future work. 
Compared to \textit{CARef}, using \textit{MaxLogit}, \textit{Energy}, and \textit{GEN} significantly improves the performance of \textit{CADRef}, while using \textit{MSP} leads to a substantial decline in its performance. 
This contrasts with the trend observed in the Table \ref{result_big_scale}, where \textit{MSP} demonstrates superior performance over both \textit{MaxLogit} and \textit{Energy}.

\subsection{Ablation Study}

\renewcommand{\arraystretch}{1.2}
\begin{table}[htbp]
\centering
\setlength{\tabcolsep}{0.9mm}
\begin{tabular}{c cc cc cc }
\Xhline{1px}
\multirow{2}{*}{\bf{Architectures}} 
& \multicolumn{2}{c}{\textit{\( \ell_1 \)-Distance}} & \multicolumn{2}{c}{\textit{\( \ell_1 \)-Norm}} & \multicolumn{2}{c}{\textit{CARef}} \\
\cline{2-7}
& AU$\uparrow$          & FP$\downarrow$         & AU$\uparrow$         & FP$\downarrow$   & AU$\uparrow$         & FP$\downarrow$        \\ \hline 
ResNet & 74.26& 78.65 &79.69& 55.20 &\bf{89.94}& \bf{40.91} \\ 
\hline
ViT & 85.16 &66.11 &20.11 &99.44& \bf{86.84}& \bf{60.48} \\
\hline
Swin & 85.83 &67.05& 15.95 &99.75& \bf{86.92} &\bf{58.65} \\
\hline
ConvNeXt & 87.49& 58.54 & 14.09 &99.83& \bf{87.95}& \bf{54.09} \\
\hline
DenseNet & 69.32 &85.54 &68.66& 75.31& \bf{86.55}& \bf{52.59}\\
\hline
RegNet & 77.30& 76.17& 65.36& 86.11& \bf{88.27}& \bf{50.68} \\
\hline
MaxViT & 85.99 &64.43& 22.12 &99.46& \bf{87.70}& \bf{50.75} \\
\hline
Average & 62.19 &80.76& 40.85& 87.87& \bf{87.74} &\bf{52.59} \\
\Xhline{1px}
\end{tabular}
\caption{Ablation Study of \textit{CARef}.
\textit{\( \ell_1 \)-Distance} represents the negative of the \( \ell_1 \) distance between the sample feature and the class-aware average feature as the score function, while \textit{\( \ell_1 \)-Norm} uses the \( \ell_1 \) norm of the sample feature as the score function.}
 \label{ablation_on_caref}
\end{table}

\noindent\textbf{Ablation of \textit{CARef}.} 
In Table \ref{ablation_on_caref}, we presents the ablation results of \textit{CARef} on the ImageNet-1k benchmark, with all values averaged across multiple OOD datasets.
Experiments show that using either \textit{\( \ell_1 \)-Distance} or \textit{\( \ell_1 \)-Norm} alone results in a significant performance gap compared to the baseline. 
Yu \textit{et al.} observe on ResNet that ID samples generally exhibit a larger feature norm than OOD samples \cite{FeatureNorm}. 
The \textit{\( \ell_1 \)-Norm} is similar to their proposed \textit{FeatureNorm}, with the main distinction being that \textit{FeatureNorm} focuses on a specific feature layer block.
However, our experiments reveal that this phenomenon does not generalize across different models.
Specifically, our experiments with models including ViT, Swin, ConvNeXt, and MaxViT demonstrate contrary behavior, with \textit{\( \ell_1 \)-Norm} achieving AUROC scores below 30\% and FPR95 values exceeding 99\%.
This suggests that the scoring for ID and OOD samples is reversed, meaning that in most cases, the feature norm of OOD samples is greater than that of ID samples.
In contrast, \textit{CARef}, as a combination of both, demonstrates significant performance improvement and robustness across multiple models.

\begin{table}[htbp]
\centering
\setlength{\tabcolsep}{1mm}
\begin{tabularx}{0.4\textwidth}{>{\hsize=0.5\hsize\centering\arraybackslash}X|>{\hsize=0.5\hsize\centering\arraybackslash}X|>{\hsize=0.5\hsize\centering\arraybackslash}X|>{\hsize=0.5\hsize\centering\arraybackslash}X|>{\hsize=0.5\hsize\centering\arraybackslash}X|>{\hsize=0.5\hsize\centering\arraybackslash}X}
\Xhline{1px}
\textbf{FD} & \textbf{$\mathcal{E}_p$}  & \textbf{$\mathcal{E}_n$} & \textbf{ES} & \textbf{AU$\uparrow$} & \textbf{FP$\downarrow$} \\
\hline
\textcolor[rgb]{0.8,0,0}{\ding{55}} &\textbf{\textemdash}&\textbf{\textemdash} &\textbf{\textemdash}&\underline{87.74} &52.59\\
\textcolor[rgb]{0,0.5,0}{\ding{51}}&\textcolor[rgb]{0,0.5,0}{\ding{51}}& \textcolor[rgb]{0.8,0,0}{\ding{55}} & \textcolor[rgb]{0.8,0,0}{\ding{55}}&82.52 & 65.67\\
\textcolor[rgb]{0,0.5,0}{\ding{51}}&\textcolor[rgb]{0.8,0,0}{\ding{55}} &\textcolor[rgb]{0,0.5,0}{\ding{51}} & \textcolor[rgb]{0.8,0,0}{\ding{55}} &87.70  & \underline{51.82}\\
\textcolor[rgb]{0,0.5,0}{\ding{51}}&\textcolor[rgb]{0,0.5,0}{\ding{51}}& \textcolor[rgb]{0.8,0,0}{\ding{55}} &\textcolor[rgb]{0,0.5,0}{\ding{51}} &87.58  & 52.43\\
\textcolor[rgb]{0,0.5,0}{\ding{51}}&\textcolor[rgb]{0,0.5,0}{\ding{51}}& \textcolor[rgb]{0,0.5,0}{\ding{51}} &  \textcolor[rgb]{0,0.5,0}{\ding{51}} &\textbf{88.19 } &  \textbf{50.30}\\
\Xhline{1px}
\end{tabularx}
\caption{Ablation Study of \textit{CADRef}.
\textbf{FD} and \textbf{ES} represent the \textit{Feature Decoupling} and \textit{Error Scaling} components, respectively.
\textcolor[rgb]{0,0.5,0}{\ding{51}} and \textcolor[rgb]{0.8,0,0}{\ding{55}} indicate whether the component is used or not.}
\label{ablation_on_cadref}
\end{table}

\noindent\textbf{Ablation of \textit{CADRef}.}
As shown in Table \ref{ablation_on_cadref}, we also conduct ablation experiments to verify the effectiveness of each module of \textit{CADRef}.
The first and fifth rows show the results for \textit{CARef} and \textit{CADRef}, respectively.
The experimental results demonstrate that removing any component results in a degradation in the performance of \textit{CADRef}, which validates that each component plays a crucial role.
The second and third rows of the table clearly show that using the negative error significantly outperforms using the positive error. 
Furthermore, the AUROC result obtained by using the negative error alone is comparable to that of \textit{CARef}, with even a lower FPR95. 
This suggests that the effect of using the positive error without \textit{Error Scaling} can be considered negligible.
However, once \textit{Error Scaling} is applied to the positive error (the fourth row), its performance becomes comparable to that of the negative error.
Note that \textit{Error Scaling} cannot be applied independently of {Feature Decoupling}, so we can only validate their collaborative effectiveness, as reflected in the performance gap between \textit{CADRef} and \textit{CARef}.

\subsection{Discussion}

\renewcommand{\arraystretch}{1.35}
\begin{table}[htbp]
\centering
\setlength{\tabcolsep}{0.95mm}
\begin{tabular}{c|cc|cc|cc }
\Xhline{1px}
\multirow{2}{*}{\bf{Methods}} 
& \multicolumn{2}{c|}{\bf{ImageNet-O}} & \multicolumn{2}{c|}{\bf{SSB-hard}} & \multicolumn{2}{c}{\bf{Ninco}} \\
\cline{2-7}
& AU$\uparrow$          & FP$\downarrow$         & AU$\uparrow$         & FP$\downarrow$   & AU$\uparrow$         & FP$\downarrow$        \\ \hline 
\textit{Energy} \cite{Energy} &50.25 &92.71& 66.99 &83.40& 72.41& 76.59 \\
\hline 

\textit{Residual} \cite{VIM} & \textbf{76.37} &\textbf{78.99}& 57.06& 89.05 &70.47 & 80.18 \\
\textit{ViM} \cite{VIM} & 74.68 &  82.14 & \textbf{68.81} &  \textbf{85.16} & \textbf{81.72} &  \textbf{71.51} \\
\hline 
\textit{CARef} &\textbf{78.03} &  \textbf{81.73} & 72.32  & 81.04 & 83.65  & 68.27 \\
\textit{CADRef} & 75.29  & 85.21
 & \textbf{74.58}  & \textbf{78.79} & \textbf{85.36 } & \textbf{64.89}
 \\
\Xhline{1px}
\end{tabular}
\caption{The effect of the logit-based scoring component on \textit{CADRef} and \textit{ViM} on three hard OOD dataset.
Both \textit{CADRef} and \textit{ViM} use \textit{Energy} as their logit-based component.}
\label{imagenet_o_performance}
\vspace{-0.15in}
\end{table}
In this subsection, we examine the limitations of \textit{CADRef}'s logit-based component through experiments on three hard OOD benchmarks: ImageNet-O \cite{imagenet-o}, SSB-hard \cite{ssb_hard}, and Ninco \cite{ninco},
which have been empirically shown to be challenging for logit-based methods.
Additionally, we also include a comparative analysis between \textit{ViM} and its feature-only component, \textit{Residual} \cite{VIM}.
As shown in Table \ref{imagenet_o_performance}, the performance of \textit{Energy} on ImageNet-O degrades to an AUROC of approximately 50\%, essentially reducing to random classification.
This degradation is reflected in both \textit{CADRef} and \textit{ViM}, which underperform their respective feature-only counterparts on this dataset.
While SSB-hard and Ninco are also considered hard OOD datasets, \textit{Energy} maintains discriminative capability with AUROC scores above 60\%. 
In these cases, \textit{CADRef} demonstrates superior performance compared to \textit{CARef}, a pattern similarly observed in the comparison between \textit{ViM} and \textit{Residual}.
These empirical findings lead to two key conclusions: (1) When logit-based methods encounter catastrophic failure on extremely challenging OOD datasets, their integration into \textit{CADRef} becomes detrimental to overall performance; (2) However, in scenarios where logit-based methods maintain even modest discriminative power, they contribute positively to the effectiveness of \textit{CADRef}.
\section{Conclusion}
\label{sec:conclusion}

In this paper, we presented a novel OOD detection framework, \textit{CADRef}, which leverages class-aware decoupled relative features to enhance the detection of out-of-distribution samples. Building on the class-aware relative error approach of \textit{CARef}, \textit{CADRef} incorporates feature decoupling and error scaling, allowing for a more nuanced separation of in-distribution and out-of-distribution samples based on their positive and negative feature contributions. Comprehensive experiments across both large-scale and small-scale benchmarks demonstrate the robustness and effectiveness of \textit{CADRef}, particularly when combined with advanced logit-based scores such as  \textit{GEN}, yielding superior AUROC and FPR95 metrics compared to state-of-the-art baselines. Future work may investigate additional decoupling strategies and adaptive scaling techniques to further enhance detection reliability across diverse datasets and architectures.

\section*{Acknowledgment}
This work was supported in part by the National Key Research and Development Program of China under Grant 2024YFB3309400, the National Science Foundation of China (62125206, 62472375, 62472338), the Major Program of the National Natural Science Foundation of Zhejiang (LD24F020014, LD25F020002), and the Zhejiang Pioneer (Jianbing) Project (2024C01032). Hailiang Zhao’s work was supported in part by the Zhejiang University Education Foundation Qizhen Scholar Foundation.
{
    \small
    \bibliographystyle{plain}
    \bibliography{main}
}

\clearpage
\setcounter{page}{1}
\maketitlesupplementary

\section{Experiment Details}
\subsection{Dataset Settings}
In our experiments, we perform OOD detection across two dataset scenarios.
In the small-scale dataset scenario, we employ CIFAR-10 and CIFAR-100 as ID datasets, with SVHN, LSUN-R, LSUN-C, iSUN, Texture, and Places serving as the corresponding OOD datasets.
In the large-scale dataset scenario, we use ImageNet-1k as the ID dataset, with OOD detection conducted on six datasets: iNaturalist, SUN, Places, Textures, OpenImage-O and ImageNet-O.
Additionally, we also conduct extended experiments on SSB-hard and Ninco to further discuss the impact of the logit-based component.
Table \ref{Dataset} provides detailed information of the datasets utilized in both scenarios.

\begin{table}[h]
\centering
\setlength{\tabcolsep}{0.9mm}
\begin{tabular}{llccc}
\toprule
\multirow{2}{*}{} &\textbf{Dataset} & \textbf{\# of Classes} & \textbf{\# of Samples} & \textbf{Size} \\
\midrule
\multirow {3}{*}{\rotatebox{90}{\textit{ID}}} &\textit{CIFAR-10} & $10$ & $10000$ & $32\times32$ \\
&\textit{CIFAR-100} & $100$ & $10000$ & $32\times32$ \\
&\textit{ImageNet-1k} & $1000$ & $50000$ & $224\times224$ \\
\midrule
\multirow {6}{*}{\rotatebox{90}{\textit{OOD (small-scale)}}} &\textit{SVHN} & -- & 26032 & $32\times32$ \\
&\textit{LSUN-R} & -- & 10000 & $32\times32$ \\
&\textit{LSUN-C} & -- & 10000 & $32\times32$ \\
&\textit{iSUN} & -- &8925  & $32\times32$ \\
&\textit{Texture} & -- &5640 & $32\times32$ \\
&\textit{Places} & -- & 10000 & $32\times32$ \\
\midrule
\multirow {8}{*}{\rotatebox{90}{\textit{OOD (large-scale)}}}&\textit{iNaturalist} & -- & 10000 & $224\times224$ \\
&\textit{SUN} & -- & 10000 & $224\times224$ \\
&\textit{Places} & -- & 10000  & $224\times224$ \\
&\textit{Texture} & -- &  5640& $224\times224$ \\
&\textit{OpenImage-O} & -- & 17632  & $224\times224$ \\
&\textit{ImageNet-O} & -- & 2000 & $224\times224$ \\
&\textit{SSB-hard} & -- & 49000 & $224\times224$ \\
&\textit{Ninco} & -- & 5878 & $224\times224$ \\
\bottomrule
\end{tabular}
\caption{
Details of ID/OOD datasets. 
Note that the third column is the number of test samples.
All OOD dataset samples are processed to match the size of the corresponding ID dataset samples.}
\label{Dataset}
\end{table}

\begin{table*}[h]
\centering
\begin{tabular}{llccc}
\toprule
\textbf{Method} & \textbf{Configurable Hyperparameters} & \textbf{CIFAR-10} & \textbf{CIFAR-100} & \textbf{ImageNet-1k} \\
\midrule
\textit{MSP} & None & -- & -- & -- \\
\midrule
\textit{MaxLogit} & None & -- & -- & -- \\
\midrule
\textit{ODIN} & $T$: temperature scaling & $T=1000$ & $T=1000$ & $T=1000$ \\
     & $\epsilon$: perturbation magnitude & $\epsilon=0.0014$ & $\epsilon=0.002$ & $\epsilon=0.005$ \\
\midrule
\textit{Energy} & $T$: temperature scaling& $T=1$ & $T=1$ & $T=1$ \\
\midrule
\textit{GEN} & $M$: top classes used in truncated sum & $M=10$ & $M=10$ & $M=100$ \\
& $\gamma$: exponential scale & $\gamma=0.1$ &  $\gamma=0.1$ &  $\gamma=0.1$ \\
\midrule
\textit{ReAct} & $p$: percentile for rectification threshold & $p=90$ & $p=90$ & $p=90$ \\
\midrule
\textit{DICE} & $p$: sparsity parameter & $p=0.9$ & $p=0.7$ & $p=0.7$ \\
\midrule
\textit{ViM} & $D$: dimension of principal space & $D=171$ & $D=171$  & \parbox[c]{4cm}{
    $D=  \begin{cases}
        256 & \text{for MaxViT} \\
        1000 & \text{for ResNet} \\ 
        512 & \text{for others}
    \end{cases} $
}
 \\
\midrule
\textit{ASH-S} & $p$: pruning percentage &  $p=95$ &$p=90$ & $p=90$ \\
\midrule
\textit{OptFS} & $K$: number of intervals& $K=100$ & $K=100$ & $K=100$ \\
\midrule
\textit{Ours} & None &-- & -- & --\\
\bottomrule
\end{tabular}
\caption{Hyperparameter settings for different OOD detection methods.}
\label{hyperparameter}
\end{table*}

\subsection{Hyperparameter Settings}

The detailed hyperparameters of all baseline methods are listed in Table \ref{hyperparameter}.
For most baseline methods, we keep the same hyperparameters as the original paper.
For methods that lack experiments on ImageNet-1k in the original work, we adopt the settings from the \textit{ReAct} \cite{ReAct}.

\section{Detailed Experiment Results}

For ImageNet-1k benchmark, we conduct extensive experiments using ResNet-50, RegNet-8GF, DenseNet-201, ViT-B/16, Swin-B, ConvNeXt-B, and MaxVit-B models to complete Table \ref{result_big_scale}.
All results are presented in Tables \ref{resnet_result}, \ref{regnet_result}, \ref{densenet_result}, \ref{vit_result}, \ref{swin_result}, \ref{convnext_result}, and \ref{maxvit_result}, where \textit{CARef} and \textit{CADRef} consistently demonstrate superior performance across these models, highlighting the robustness of our methods.

\section{Collapse of the feature norm}
\begin{figure}[h]
    \centering
    \begin{subfigure}{0.15\textwidth}
        \centering
        \includegraphics[width=\linewidth]{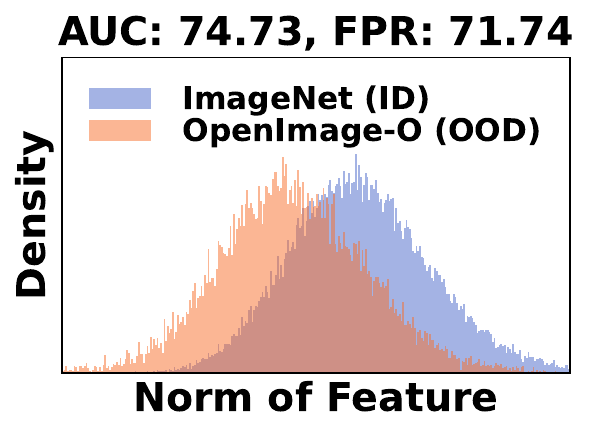}
        \caption{{ResNet}}
        \label{fig:norm_resnet}
    \end{subfigure}
    \hfill
    \begin{subfigure}{0.15\textwidth}
        \centering
        \includegraphics[width=\linewidth]{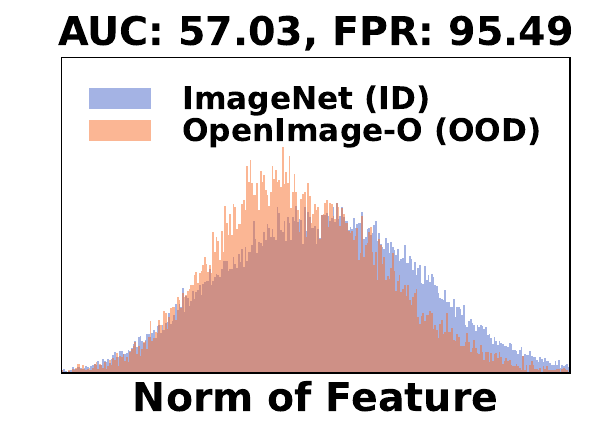}
        \caption{{RegNet}}
        \label{fig:norm_regnet}
    \end{subfigure}
    \hfill
     \begin{subfigure}{0.15\textwidth}
        \centering
        \includegraphics[width=\linewidth]{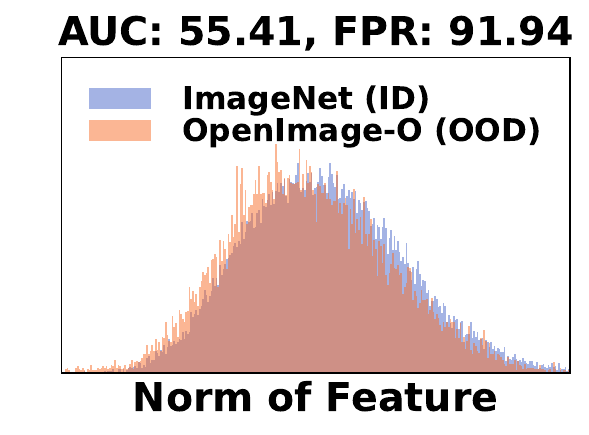}
        \caption{{DenseNet}}
        \label{fig:norm_densenet}
    \end{subfigure}
     \hfill
     \begin{subfigure}{0.15\textwidth}
        \centering
        \includegraphics[width=\linewidth]{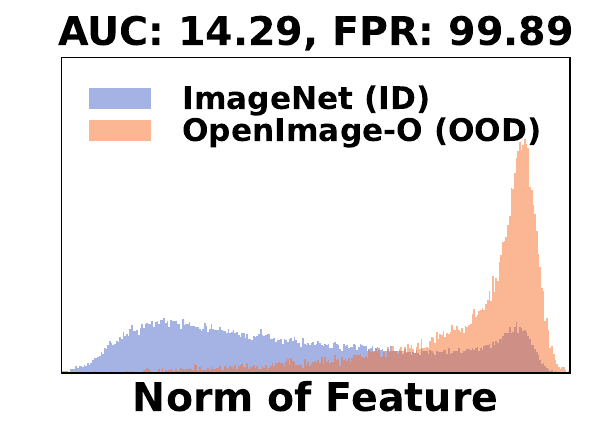}
        \caption{{ViT}}
        \label{fig:norm_vit}
    \end{subfigure}
     \hfill
     \begin{subfigure}{0.15\textwidth}
        \centering
        \includegraphics[width=\linewidth]{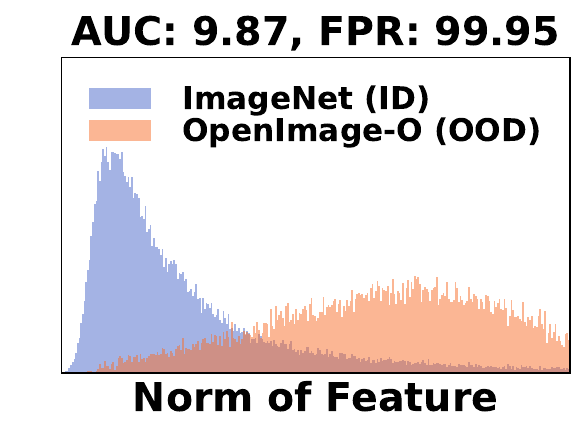}
        \caption{{Swin}}
        \label{fig:norm_swin}
    \end{subfigure}
     \hfill
     \begin{subfigure}{0.15\textwidth}
        \centering
        \includegraphics[width=\linewidth]{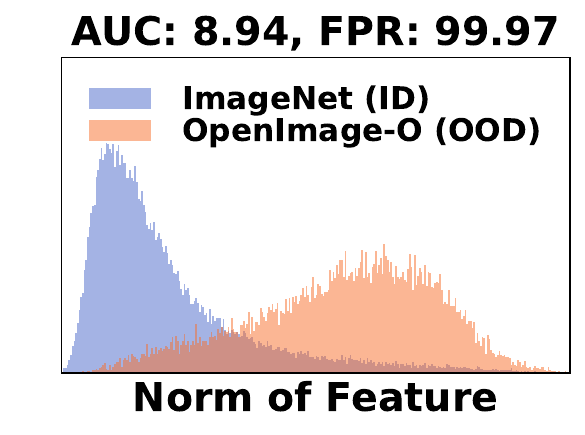}
        \caption{{ConvNeXt}}
        \label{fig:norm_convnext}
    \end{subfigure}
     \hfill
     \begin{subfigure}{0.15\textwidth}
        \centering
        \includegraphics[width=\linewidth]{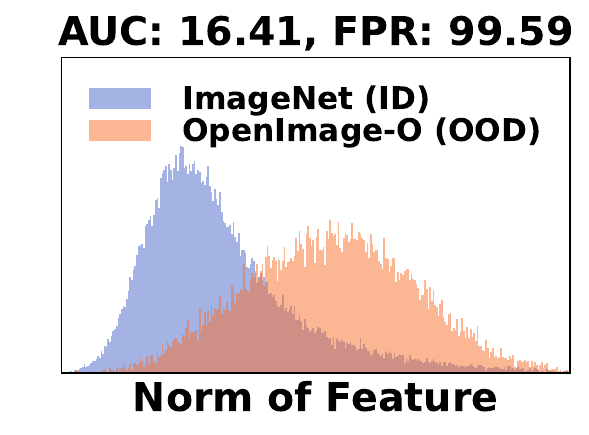}
        \caption{{MaxViT}}
        \label{fig:norm_maxvit}
    \end{subfigure}
    \caption{Comparison of feature \( \ell_2 \)-norm across different models.}
    \label{fig:norm}
\end{figure}
We also explore the challenges associated with using feature norm for OOD detection.
According to the conclusion in \cite{FeatureNorm}, the feature norms of ID samples are generally larger than those of OOD samples.
However, as shown in Figure \ref{fig:norm}, this conclusion holds only for ResNet. 
In RegNet and DenseNet, this method lacks discriminative power (with an AUROC of around 50\%), and in other models, the conclusion is actually reversed.

\subsection{Score and error distribution for other methods}

\begin{figure}[htbp]
    \centering
    \begin{subfigure}{0.235\textwidth}
        \centering
        \includegraphics[width=\linewidth]{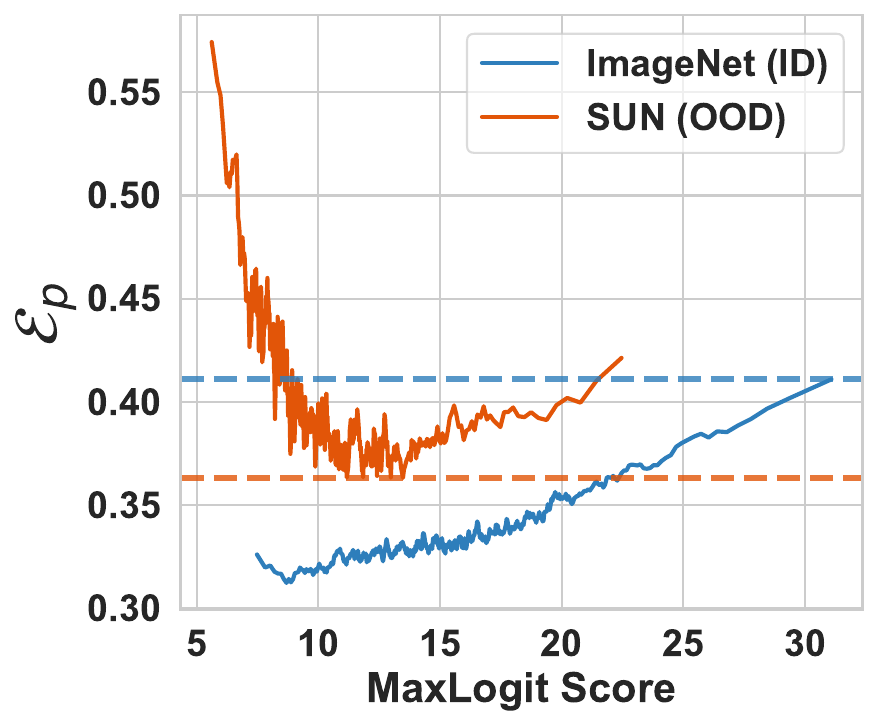}
        \caption{Positive Feature}
        \label{fig:maxlogit_p}
    \end{subfigure}
    \hfill
    \begin{subfigure}{0.235\textwidth}
        \centering
        \includegraphics[width=\linewidth]{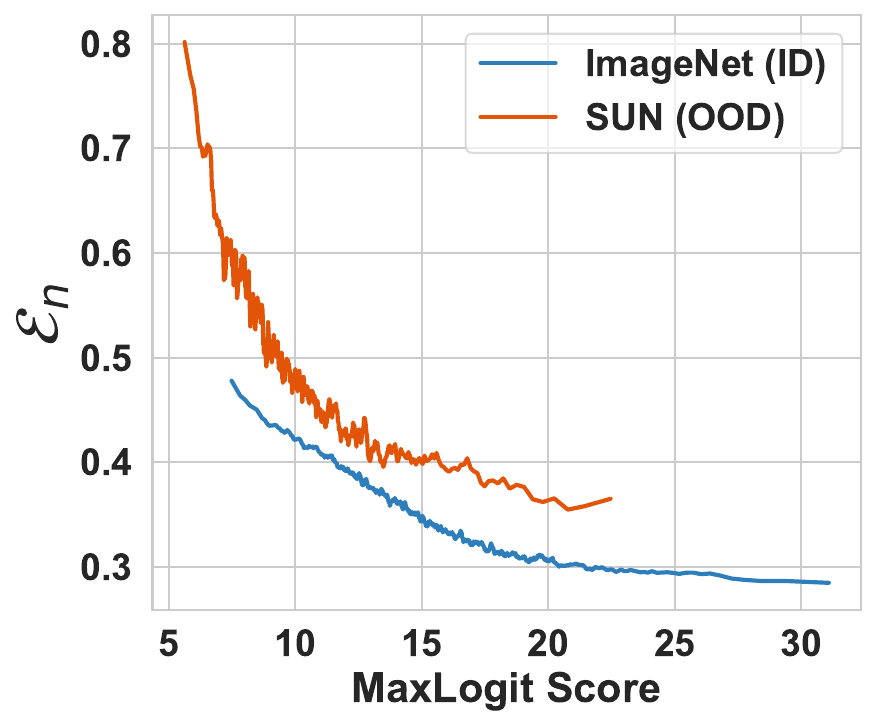}
        \caption{Negative Feature}
        \label{fig:maxlogit_n}
    \end{subfigure}
    \\
    \begin{subfigure}{0.235\textwidth}
        \centering
        \includegraphics[width=\linewidth]{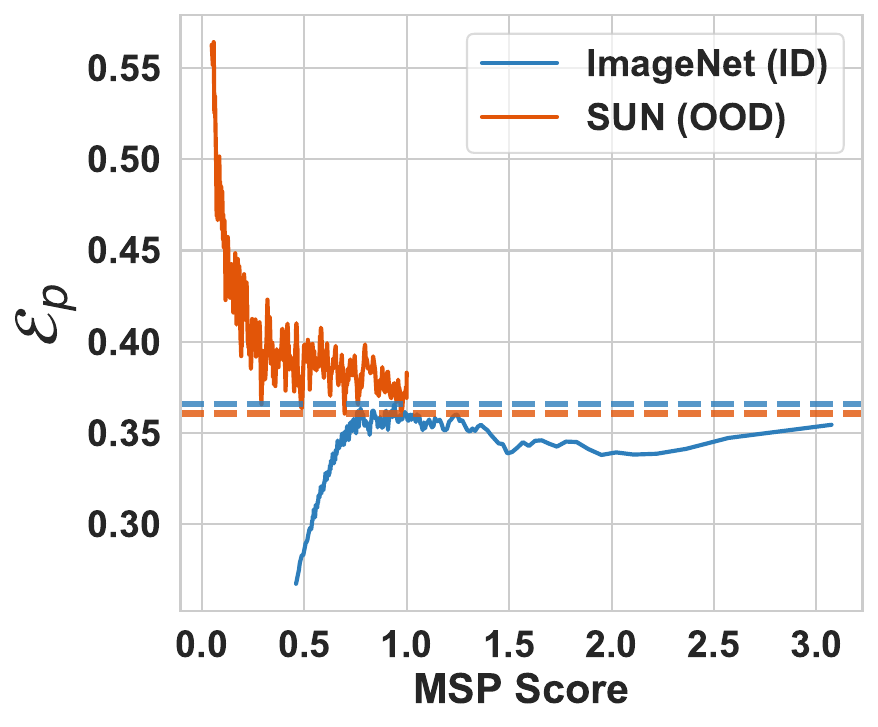}
        \caption{Positive Feature}
        \label{fig:msp_p}
    \end{subfigure}
    \hfill
    \begin{subfigure}{0.235\textwidth}
        \centering
        \includegraphics[width=\linewidth]{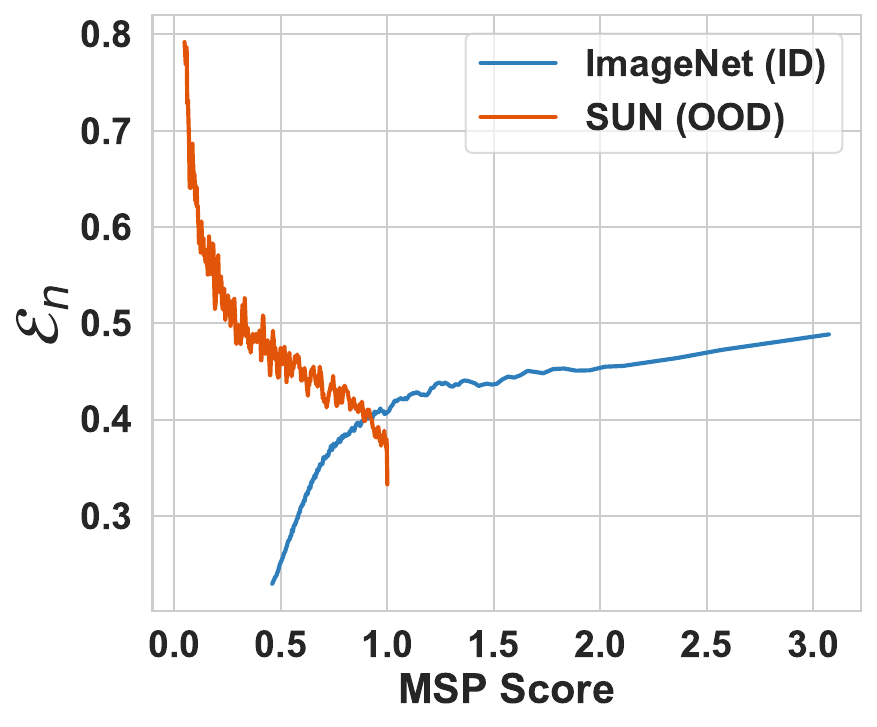}
        \caption{Negative Feature}
        \label{fig:msp_n}
    \end{subfigure}
    \caption{Score and error distribution of ID/OOD samples.}
    \label{supp}
\end{figure}

Figure \ref{supp} illustrates the score and error distributions for the \textit{MSP} and \textit{MaxLogit} methods. While the \textit{MaxLogit} method demonstrates error trends consistent with the \textit{Energy} and \textit{GEN} methods, the \textit{MSP} method exhibits a markedly different behavior. At higher $\mathcal{S}_\text{logit}$ values, \textit{MSP} shows distinctly separated positive errors but significant overlap in negative errors, which fundamentally contradicts the \textit{Error Scaling} component's design principles. Consequently, integrating \textit{MSP} as a logit-based component in \textit{CADRef} leads to a detrimental impact, as substantiated by the results in Figure \ref{impact}.

\begin{table*}[tbp]
\centering
\setlength{\tabcolsep}{1.3mm}
\begin{NiceTabular}{c |cc| cc| cc |cc | cc | cc |cc}
\CodeBefore
\rectanglecolor[gray]{0.9}{1-14}{14-15}
\Body
\Xhline{1px}
\multirow{2}{*}{\bf{Methods}} 
& \multicolumn{2}{c|}{\bf{iNaturalist}} & \multicolumn{2}{c|}{\bf{SUN}} & \multicolumn{2}{c|}{\bf{Places}} & \multicolumn{2}{c|}{\bf{Textures}} & \multicolumn{2}{c|}{\bf{OpenImage-O}} & \multicolumn{2}{c|}{\bf{ImageNet-O}}   &  \multicolumn{2}{c}{\cellcolor[gray]{0.9}\bf{Average}}  \\ 
\hhline{~*{12}{-}|*{2}{-}}
& AU$\uparrow$          & FP$\downarrow$         & AU$\uparrow$         & FP$\downarrow$   & AU$\uparrow$         & FP$\downarrow$      & AU$\uparrow$           & FP$\downarrow$ & AU$\uparrow$          & FP$\downarrow$
                          & AU$\uparrow$           & FP$\downarrow$& AU$\uparrow$           & FP$\downarrow$ \\ \hline 
\textit{MSP} & 88.42 & 52.73 & 81.75 & 68.58 & 80.63 & 71.59 & 80.46 & 66.15 & 84.98 & 63.60 & 28.61 & 100.00 & 74.14 & 70.44 \\
\textit{MaxLogit} & 91.14 & 50.77 & 86.43 & 60.39 & 84.03 & 66.03 & 86.38 & 54.91 & 89.13 & 57.89 & 40.73 & 100.00 & 79.64 & 65.00 \\
\textit{ODIN} & 87.00 & 52.33 & 86.57 & 53.49 & 85.30 & 58.64 & 86.51 & 46.08 & 86.65 & 52.76 & 43.63 & 98.65 & 79.28 & 60.32 \\
\textit{Energy} & 90.59 & 53.96 & 86.73 & 58.28 & 84.12 & 65.43 & 86.73 & 52.30 & 89.12 & 57.23 & 41.79 & 100.00 & 79.85 & 64.53 \\
\textit{GEN} & 92.44 & 45.76 & 85.52 & 65.54 & 83.46 & 69.24 & 85.41 & 59.24 & 89.31 & 60.44 & 43.59 & 100.00 & 79.95 & 66.70 \\
\textit{ReAct} & 96.39 & 19.55 & \textbf{94.41} & \textbf{24.01} & \textbf{91.93} & \textbf{33.45} & 90.45 & 45.83 & 90.53 & 43.69 & 52.45 & 98.00 & 86.03 & 44.09 \\
\textit{DICE} & 94.51 & 26.63 & 90.91 & 36.48 & 87.64 & 47.98 & 90.44 & 32.58 & 88.57 & 45.72 & 42.78 & 98.00 & 82.48 & 47.90 \\
\textit{ViM} & 87.42 & 71.80 & 81.07 & 81.80 & 78.39 & 83.12 & 96.83 & 14.84 & 89.30 & 58.68 & \underline{70.77} & \underline{84.85} & 83.96 & 65.85 \\
\textit{ASH-S} & \textbf{97.87} & \textbf{11.49} & \underline{94.02} & \underline{27.96} & \underline{90.98} & \underline{39.83} & \underline{97.60} & \underline{11.97} & 92.75 & \underline{32.77} & 67.44 & 89.10 & \textbf{90.11} & \textbf{35.52} \\
\textit{OptFS} & 96.88 & 16.79 & 93.13 & 35.31 & 90.42 & 44.78 & 95.74 & 23.08 & \underline{92.77} & 37.68 & 59.94 & 97.20 & 88.15 & 42.47 \\
\hline
\textit{CARef} & 96.54 & 17.46 & 89.51 & 44.89 & 85.41 & 57.64 & \textbf{97.94} & \textbf{10.15} & 92.57 & 37.73 & \textbf{77.66} & \textbf{77.60} & \underline{89.94} & 40.91 \\
\textit{CADRef} & \underline{96.90} & \underline{16.08} & 91.26 & 39.23 & 87.80 & 51.12 & 97.14 & 12.60 & \textbf{93.93} & \textbf{32.69} & 68.38 & 92.35 & 89.24 & \underline{40.68} \\
\Xhline{1px}
\end{NiceTabular}
\caption{Results of OOD detection on ImageNet-1k benchmark with ResNet-50. 
$\uparrow$ indicates that higher values are better, while $\downarrow$ indicates that lower values are better.
All values are percentages, with the best and second-best results being \textbf{highlighted} and \underline{underlined}, respectively.}
\label{resnet_result}
\end{table*}

\begin{table*}[htbp]
\centering
\setlength{\tabcolsep}{1.3mm}
\begin{NiceTabular}{c |cc| cc| cc |cc | cc | cc |cc}
\CodeBefore
\rectanglecolor[gray]{0.9}{1-14}{14-15}
\Body
\Xhline{1px}
\multirow{2}{*}{\bf{Methods}} 
& \multicolumn{2}{c|}{\bf{iNaturalist}} & \multicolumn{2}{c|}{\bf{SUN}} & \multicolumn{2}{c|}{\bf{Places}} & \multicolumn{2}{c|}{\bf{Textures}} & \multicolumn{2}{c|}{\bf{OpenImage-O}} & \multicolumn{2}{c|}{\bf{ImageNet-O}}   &  \multicolumn{2}{c}{\cellcolor[gray]{0.9}\bf{Average}}  \\ 
\hhline{~*{12}{-}|*{2}{-}}
& AU$\uparrow$          & FP$\downarrow$         & AU$\uparrow$         & FP$\downarrow$   & AU$\uparrow$         & FP$\downarrow$      & AU$\uparrow$           & FP$\downarrow$ & AU$\uparrow$          & FP$\downarrow$
                          & AU$\uparrow$           & FP$\downarrow$ & AU$\uparrow$           & FP$\downarrow$ \\ \hline 
\textit{MSP} & 88.32 & 53.48 & 82.16 & 68.00 & 80.98 & 71.68 & 79.87 & 68.77 & 86.01 & 61.30 & 52.69 & 96.65 & 78.34 & 69.98 \\
\textit{MaxLogit} & 90.01 & 51.89 & 86.32 & 57.56 & 84.02 & 64.45 & 82.71 & 61.24 & 89.59 & 51.53 & 57.64 & 95.80 & 81.71 & 63.75 \\
\textit{ODIN} & 85.52 & 54.58 & 86.11 & 50.33 & 85.76 & 54.66 & 82.34 & 53.19 & 86.99 & 49.30 & 56.72 & 89.90 & 80.57 & 58.66 \\
\textit{Energy} & 89.29 & 55.36 & 86.27 & 57.37 & 83.85 & 63.97 & 82.51 & 61.40 & 89.36 & 51.87 & 58.03 & 94.70 & 81.55 & 64.11 \\
\textit{GEN} & 92.36 & 44.63 & 86.44 & 60.33 & 84.40 & 66.31 & 84.53 & 60.85 & 90.59 & 52.96 & 62.12 & 95.65 & 83.41 & 63.46 \\
\textit{ReAct} & 96.04 & 21.72 & \textbf{94.85} & \textbf{24.55} & \textbf{91.71} & \textbf{36.38} & 87.26 & 56.80 & 87.96 & 47.03 & 61.91 & 91.35 & 86.62 & 46.30 \\
\textit{DICE} & 88.83 & 56.48 & 84.19 & 62.09 & 80.05 & 74.33 & 80.13 & 64.18 & 81.65 & 62.26 & 52.65 & 94.00 & 77.92 & 68.89 \\
\textit{ViM} & 90.88 & 58.04 & 85.31 & 70.72 & 82.12 & 75.84 & \textbf{97.35} & \textbf{12.23} & 91.95 & 48.58 & \underline{80.85} & \underline{71.40} & 88.08 & 56.13 \\
\textit{ASH-S} & \textbf{96.49} & \textbf{18.52} & 91.00 & \underline{35.15} & 86.84 & \underline{50.60} & 97.12 & \underline{13.40} & 91.07 & \underline{35.58} & 73.71 & 74.80 & \underline{89.37} & \textbf{38.01} \\
\textit{OptFS} & \underline{96.06} & \underline{20.96} & \underline{92.03} & 40.58 & \underline{88.30} & 51.62 & 95.90 & 22.96 & \underline{92.33} & 37.47 & 71.19 & 84.20 & 89.30 & 42.97 \\
\hline
\textit{CARef} & 93.07 & 41.97 & 86.21 & 57.28 & 81.48 & 72.04 & \underline{97.26} & 15.00 & 90.21 & 47.77 & \textbf{81.38} & \textbf{70.00} & 88.27 & 50.68 \\
\textit{CADRef} & 95.31 & 28.13 & 90.29 & 44.08 & 86.73 & 58.14 & 97.20 & 14.02 & \textbf{93.87} & \textbf{34.70} & 78.15 & 75.00 & \textbf{90.26} & \underline{42.34} \\
\Xhline{1px}
\end{NiceTabular}
\caption{Results of OOD detection on ImageNet-1k benchmark with RegNet-8GF. 
$\uparrow$ indicates that higher values are better, while $\downarrow$ indicates that lower values are better.
All values are percentages, with the best and second-best results being \textbf{highlighted} and \underline{underlined}, respectively.}
\label{regnet_result}
\end{table*}

\begin{table*}[htbp]
\centering
\setlength{\tabcolsep}{1.3mm}
\begin{NiceTabular}{c |cc| cc| cc |cc | cc | cc |cc}
\CodeBefore
\rectanglecolor[gray]{0.9}{1-14}{14-15}
\Body
\Xhline{1px}
\multirow{2}{*}{\bf{Methods}} 
& \multicolumn{2}{c|}{\bf{iNaturalist}} & \multicolumn{2}{c|}{\bf{SUN}} & \multicolumn{2}{c|}{\bf{Places}} & \multicolumn{2}{c|}{\bf{Textures}} & \multicolumn{2}{c|}{\bf{OpenImage-O}} & \multicolumn{2}{c|}{\bf{ImageNet-O}}   &  \multicolumn{2}{c}{\cellcolor[gray]{0.9}\bf{Average}}  \\ 
\hhline{~*{12}{-}|*{2}{-}}
& AU$\uparrow$          & FP$\downarrow$         & AU$\uparrow$         & FP$\downarrow$   & AU$\uparrow$         & FP$\downarrow$      & AU$\uparrow$           & FP$\downarrow$ & AU$\uparrow$          & FP$\downarrow$
                          & AU$\uparrow$           & FP$\downarrow$ & AU$\uparrow$           & FP$\downarrow$ \\ \hline 
\textit{MSP} & 89.83 & 44.80 & 82.22 & 65.22 & 81.13 & 68.86 & 79.40 & 67.02 & 85.21 & 61.93 & 48.78 & 97.95 & 77.76 & 67.63 \\
\textit{MaxLogit} & 92.12 & 40.68 & 85.91 & 55.85 & 83.83 & 62.12 & 83.41 & 58.31 & 88.31 & 55.05 & 53.35 & 97.35 & 81.16 & 61.56 \\
\textit{ODIN} & 89.95 & 42.93 & 84.09 & 55.57 & 83.01 & 59.93 & 81.45 & 54.85 & 84.70 & 56.80 & 50.21 & 94.40 & 78.90 & 60.75 \\
\textit{Energy} & 91.47 & 43.76 & 85.97 & 53.85 & 83.68 & 61.42 & 83.44 & 56.61 & 87.95 & 55.92 & 54.04 & 95.95 & 81.09 & 61.25 \\
\textit{GEN} & 93.37 & 37.92 & 85.77 & 59.34 & 84.03 & 64.53 & 83.78 & 60.69 & 89.24 & 58.20 & 58.18 & 97.40 & 82.39 & 63.01 \\
\textit{ReAct} & 88.65 & 51.99 & 89.55 & 51.38 & 85.92 & 61.57 & 81.49 & 60.83 & 74.19 & 68.20 & 50.97 & 91.10 & 78.46 & 64.18 \\
\textit{DICE} & 91.73 & 39.18 & 86.76 & 48.13 & 82.29 & 60.72 & 83.85 & 53.28 & 81.62 & 59.35 & 48.94 & 93.60 & 79.20 & 59.04 \\
\textit{ViM} & 82.69 & 85.18 & 73.99 & 91.50 & 72.87 & 91.43 & 94.66 & 26.13 & 87.21 & 66.06 & \underline{76.58} & \underline{75.64} & 81.33 & 72.66 \\
\textit{ASH-S} & \textbf{96.18} & \textbf{20.51} & \textbf{90.58} & \textbf{39.42} & \textbf{87.54} & \textbf{50.40} & 93.81 & 26.13 & \textbf{92.28} & \textbf{37.42} & 64.97 & 88.95 & 87.56 & \textbf{43.80} \\
\textit{OptFS} & 94.61 & 28.62 & \underline{90.08} & \underline{46.12} & \underline{86.18} & \underline{56.28} & 95.17 & 26.32 & 90.23 & 46.91 & 73.21 & 83.00 & \underline{88.25} & 47.88 \\
\hline
\textit{CARef} & 92.94 & 36.18 & 84.16 & 60.12 & 78.68 & 72.13 & \textbf{96.43} & \underline{18.99} & 87.01 & 54.50 & \textbf{80.08} & \textbf{73.60} & 86.55 & 52.59 \\
\textit{CADRef} & \underline{95.41} & \underline{24.59} & 89.17 & 47.19 & 85.03 & 58.96 & \underline{96.39} & \textbf{16.68} & \underline{91.57} & \underline{42.85} & 74.27 & 81.20 & \textbf{88.64} & \underline{45.25} \\
\Xhline{1px}
\end{NiceTabular}
\caption{Results of OOD detection on ImageNet-1k benchmark with DenseNet-201. 
$\uparrow$ indicates that higher values are better, while $\downarrow$ indicates that lower values are better.
All values are percentages, with the best and second-best results being \textbf{highlighted} and \underline{underlined}, respectively.}
\label{densenet_result}
\end{table*}

\begin{table*}[htbp]
\centering
\setlength{\tabcolsep}{1.3mm}
\begin{NiceTabular}{c |cc| cc| cc |cc | cc | cc |cc}
\CodeBefore
\rectanglecolor[gray]{0.9}{1-14}{14-15}
\Body
\Xhline{1px}
\multirow{2}{*}{\bf{Methods}} 
& \multicolumn{2}{c|}{\bf{iNaturalist}} & \multicolumn{2}{c|}{\bf{SUN}} & \multicolumn{2}{c|}{\bf{Places}} & \multicolumn{2}{c|}{\bf{Textures}} & \multicolumn{2}{c|}{\bf{OpenImage-O}} & \multicolumn{2}{c|}{\bf{ImageNet-O}}   &  \multicolumn{2}{c}{\cellcolor[gray]{0.9}\bf{Average}}  \\ 
\hhline{~*{12}{-}|*{2}{-}}
& AU$\uparrow$          & FP$\downarrow$         & AU$\uparrow$         & FP$\downarrow$   & AU$\uparrow$         & FP$\downarrow$      & AU$\uparrow$           & FP$\downarrow$ & AU$\uparrow$          & FP$\downarrow$
                          & AU$\uparrow$           & FP$\downarrow$ & AU$\uparrow$           & FP$\downarrow$ \\ \hline 
\textit{MSP} & 88.18 & 51.42 & 80.88 & 66.65 & 80.37 & 68.49 & 82.96 & 60.47 & 84.81 & 59.84 & 58.79 & 90.85 & 79.33 & 66.29 \\
\textit{MaxLogit} & 85.27 & 52.14 & 76.28 & 66.86 & 75.03 & 69.19 & 81.64 & 56.57 & 81.53 & 58.72 & 54.31 & 88.94 & 75.68 & 65.40 \\
\textit{ODIN} & 71.50 & 94.86 & 56.56 & 95.31 & 56.12 & 95.49 & 66.32 & 92.18 & 62.79 & 94.79 & 55.43 & 93.35 & 61.45 & 94.33 \\
\textit{Energy} & 79.27 & 63.93 & 70.16 & 72.81 & 68.39 & 74.35 & 79.24 & 58.45 & 76.44 & 64.83 & 52.71 & 87.00 & 71.04 & 70.23 \\
\textit{GEN} & 92.91 & 40.05 & \underline{85.03} & \underline{61.06} & \textbf{83.49} & \underline{64.22} & 87.97 & 50.69 & 89.36 & 52.53 & 67.44 & 88.90 & 84.37 & \underline{59.58} \\
\textit{ReAct} & 85.99 & 65.07 & 78.93 & 72.38 & 77.48 & 73.84 & 84.62 & 57.09 & 84.21 & 64.58 & 66.69 & 87.00 & 79.65 & 69.99 \\
\textit{DICE} & 74.54 & 90.41 & 65.23 & 94.51 & 64.79 & 93.05 & 77.34 & 84.04 & 77.28 & 83.01 & 70.75 & \underline{86.05} & 71.65 & 88.51 \\
\textit{ViM} & \textbf{97.18} & \textbf{12.55} & 83.99 & \textbf{56.77} & 81.48 & \textbf{58.98} & 88.92 & \textbf{46.48} & \textbf{92.44} & \textbf{38.11} & 76.29 & \textbf{83.25} & 86.72 & \textbf{49.36} \\
\textit{ASH-S} & 6.68 & 99.99 & 16.70 & 99.73 & 18.32 & 99.65 & 24.13 & 99.13 & 14.04 & 99.84 & 28.46 & 99.40 & 18.06 & 99.62 \\
\textit{OptFS} & 89.94 & 55.59 & 84.19 & 66.36 & 82.70 & 68.39 & 86.46 & 56.60 & 88.20 & 59.86 & 71.85 & 89.15 & 83.89 & 65.99 \\
\hline
\textit{CARef} & \underline{94.07} & \underline{36.59} & 84.83 & 66.92 & 83.03 & 68.36 & \underline{89.28} & 52.57 & \underline{91.54} & \underline{50.32} & \textbf{78.31} & 88.14 & \underline{86.84} & 60.48 \\
\textit{CADRef} & 93.81 & 38.45 & \textbf{85.12} & 65.85 & \underline{83.28} & 67.91 & \textbf{89.59} & \underline{49.78} & 91.52 & 50.44 & \underline{78.12} & 87.85 & \textbf{86.91} & 60.05 \\
\Xhline{1px}
\end{NiceTabular}
\caption{Results of OOD detection on ImageNet-1k benchmark with ViT-B/16. 
$\uparrow$ indicates that higher values are better, while $\downarrow$ indicates that lower values are better.
All values are percentages, with the best and second-best results being \textbf{highlighted} and \underline{underlined}, respectively.}
\label{vit_result}
\end{table*}

\begin{table*}[htbp]
\centering
\setlength{\tabcolsep}{1.3mm}
\begin{NiceTabular}{c |cc| cc| cc |cc | cc | cc |cc}
\CodeBefore
\rectanglecolor[gray]{0.9}{1-14}{14-15}
\Body
\Xhline{1px}
\multirow{2}{*}{\bf{Methods}} 
& \multicolumn{2}{c|}{\bf{iNaturalist}} & \multicolumn{2}{c|}{\bf{SUN}} & \multicolumn{2}{c|}{\bf{Places}} & \multicolumn{2}{c|}{\bf{Textures}} & \multicolumn{2}{c|}{\bf{OpenImage-O}} & \multicolumn{2}{c|}{\bf{ImageNet-O}}   &  \multicolumn{2}{c}{\cellcolor[gray]{0.9}\bf{Average}}  \\ 
\hhline{~*{12}{-}|*{2}{-}}
& AU$\uparrow$          & FP$\downarrow$         & AU$\uparrow$         & FP$\downarrow$   & AU$\uparrow$         & FP$\downarrow$      & AU$\uparrow$           & FP$\downarrow$ & AU$\uparrow$          & FP$\downarrow$
                          & AU$\uparrow$           & FP$\downarrow$& AU$\uparrow$           & FP$\downarrow$ \\ \hline 
\textit{MSP} & 87.00 & 51.68 & 80.53 & 66.52 & 80.69 & 67.79 & 78.46 & 66.20 & 81.78 & 62.13 & 58.38 & 88.40 & 77.81 & 67.12 \\
\textit{MaxLogit} & 79.86 & 55.48 & 72.23 & 67.86 & 71.97 & 69.54 & 73.98 & 61.93 & 71.08 & 65.52 & 56.00 & 85.75 & 70.85 & 67.68 \\
\textit{ODIN} & 58.88 & 90.31 & 51.06 & 92.18 & 48.57 & 93.24 & 64.36 & 86.61 & 52.42 & 92.10 & 56.00 & 91.70 & 55.22 & 91.02 \\
\textit{Energy} & 68.57 & 72.55 & 63.08 & 78.54 & 62.48 & 78.92 & 69.58 & 65.95 & 60.23 & 76.13 & 55.13 & 83.65 & 63.18 & 75.96 \\
\textit{GEN} & 92.69 & \textbf{32.94} & \underline{85.02} & \textbf{56.61} & \textbf{84.06} & \textbf{59.95} & 85.61 & 49.93 & 87.18 & 48.66 & 67.77 & 83.90 & 83.72 & \textbf{55.33} \\
\textit{ReAct} & 88.72 & 59.29 & 81.46 & 69.10 & 80.87 & 70.79 & 84.07 & 57.83 & 85.90 & 61.58 & 70.45 & \textbf{80.85} & 81.91 & 66.57 \\
\textit{DICE} & 23.13 & 97.70 & 42.48 & 91.57 & 33.91 & 94.74 & 73.15 & 63.22 & 45.55 & 86.66 & 56.50 & 86.65 & 45.79 & 86.76 \\
\textit{ViM} & \underline{93.60} & 44.56 & 80.12 & 71.10 & 77.96 & 71.82 & 83.96 & 64.64 & 92.10 & \underline{46.36} & 76.03 & 83.80 & 83.96 & 63.63 \\
\textit{ASH-S} & 10.69 & 99.81 & 20.18 & 99.36 & 21.37 & 99.59 & 18.41 & 98.65 & 11.94 & 99.84 & 32.52 & 98.85 & 19.18 & 99.35 \\
\textit{OptFS} & 90.71 & 54.98 & 84.86 & 67.93 & \underline{83.94} & 68.63 & 85.10 & 61.68 & 90.34 & 51.19 & 73.28 & 85.75 & 84.71 & 65.03 \\
\hline
\textit{CARef} & 93.57 & 39.78 & 84.83 & 66.72 & 83.25 & 68.68 & \underline{88.74} & \underline{49.86} & \textbf{92.55} & \textbf{43.52} & \underline{78.57} & 83.35 & \underline{86.92} & 58.65 \\
\textit{CADRef} & \textbf{93.77} & \underline{37.87} & \textbf{85.12} & \underline{64.23} & 83.57 & \underline{66.71} & \textbf{89.08} & \textbf{47.38} & \underline{92.32} & 46.63 & \textbf{78.76} & \underline{81.45} & \textbf{87.10} & \underline{57.38} \\
\Xhline{1px}
\end{NiceTabular}
\caption{Results of OOD detection on ImageNet-1k benchmark with Swin-B. 
$\uparrow$ indicates that higher values are better, while $\downarrow$ indicates that lower values are better.
All values are percentages, with the best and second-best results being \textbf{highlighted} and \underline{underlined}, respectively.}
\label{swin_result}
\end{table*}

\begin{table*}[htbp]
\centering
\setlength{\tabcolsep}{1.3mm}
\begin{NiceTabular}{c |cc| cc| cc |cc | cc | cc |cc}
\CodeBefore
\rectanglecolor[gray]{0.9}{1-14}{14-15}
\Body
\Xhline{1px}
\multirow{2}{*}{\bf{Methods}} 
& \multicolumn{2}{c|}{\bf{iNaturalist}} & \multicolumn{2}{c|}{\bf{SUN}} & \multicolumn{2}{c|}{\bf{Places}} & \multicolumn{2}{c|}{\bf{Textures}} & \multicolumn{2}{c|}{\bf{OpenImage-O}} & \multicolumn{2}{c|}{\bf{ImageNet-O}}   &  \multicolumn{2}{c}{\cellcolor[gray]{0.9}\bf{Average}}  \\ 
\hhline{~*{12}{-}|*{2}{-}}
& AU$\uparrow$          & FP$\downarrow$         & AU$\uparrow$         & FP$\downarrow$   & AU$\uparrow$         & FP$\downarrow$      & AU$\uparrow$           & FP$\downarrow$ & AU$\uparrow$          & FP$\downarrow$
                          & AU$\uparrow$           & FP$\downarrow$ & AU$\uparrow$           & FP$\downarrow$ \\ \hline 
\textit{MSP} & 89.69 & 44.35 & 79.28 & 63.37 & 78.41 & 67.02 & 73.77 & 65.76 & 83.40 & 58.48 & 52.85 & 92.45 & 76.23 & 65.24 \\
\textit{MaxLogit} & 83.32 & 57.91 & 69.83 & 72.02 & 68.44 & 74.92 & 65.72 & 70.88 & 74.08 & 68.10 & 51.18 & 92.10 & 68.76 & 72.66 \\
\textit{ODIN} & 63.62 & 81.28 & 47.54 & 91.28 & 47.73 & 90.28 & 43.43 & 93.49 & 52.75 & 88.20 & 51.48 & 91.45 & 51.09 & 89.33 \\
\textit{Energy} & 57.94 & 90.10 & 51.17 & 94.53 & 50.85 & 92.84 & 52.87 & 87.26 & 54.62 & 88.78 & 50.00 & 93.00 & 52.91 & 91.09 \\
\textit{GEN} & \textbf{94.86} & \textbf{25.69} & 84.78 & \textbf{52.70} & 83.04 & \textbf{59.18} & 80.10 & 55.21 & 89.37 & 46.11 & 64.02 & 90.50 & 82.69 & \underline{54.90} \\
\textit{ReAct} & 87.94 & 68.80 & 78.51 & 78.02 & 77.07 & 79.42 & 76.45 & 71.57 & 83.30 & 73.30 & 65.61 & 90.10 & 78.15 & 76.87 \\
\textit{DICE} & 20.82 & 98.55 & 41.27 & 93.95 & 34.27 & 96.24 & 71.72 & 69.92 & 47.62 & 88.50 & 54.99 & 91.45 & 45.12 & 89.77 \\

\textit{ViM} & 93.63 & 40.84 & 85.00 & 59.57 & 82.17 & \underline{61.76} & 86.79 & 52.41 & 91.36 & 45.33 & 66.46 & 91.50 & 84.24 & 58.57 \\
\textit{ASH-S} & 3.79 & 99.91 & 10.77 & 99.56 & 14.87 & 99.02 & 33.22 & 96.20 & 17.33 & 98.67 & 41.80 & 96.00 & 20.30 & 98.23 \\
\textit{OptFS} & 92.20 & 46.02 & 85.96 & 61.69 & \underline{85.10} & 64.02 & 86.11 & 54.32 & 91.02 & 49.68 & 69.97 & 91.30 & 85.06 & 61.17 \\
\hline
\textit{CARef} & \underline{94.81} & \underline{29.79} & \textbf{87.37} & \underline{58.04} & \textbf{85.40} & 63.09 & \textbf{90.04} & \textbf{45.09} & \textbf{93.39} & \textbf{38.82} & \textbf{76.70} & \underline{89.70} & \textbf{87.95} & \textbf{54.09} \\
\textit{CADRef} & 94.50 & 33.08 & \underline{86.84} & 60.16 & 84.92 & 65.09 & \underline{89.27} & \underline{49.17} & \underline{92.85} & \underline{44.85} & \underline{76.49} & \textbf{89.15} & \underline{87.48} & 56.92 \\
\Xhline{1px}
\end{NiceTabular}
\caption{Results of OOD detection on ImageNet-1k benchmark with ConvNeXt-B. 
$\uparrow$ indicates that higher values are better, while $\downarrow$ indicates that lower values are better.
All values are percentages, with the best and second-best results being \textbf{highlighted} and \underline{underlined}, respectively.}
\label{convnext_result}
\end{table*}

\begin{table*}[htbp]
\centering
\setlength{\tabcolsep}{1.3mm}
\begin{NiceTabular}{c |cc| cc| cc |cc | cc | cc |cc}
\CodeBefore
\rectanglecolor[gray]{0.9}{1-14}{14-15}
\Body
\Xhline{1px}
\multirow{2}{*}{\bf{Methods}} 
& \multicolumn{2}{c|}{\bf{iNaturalist}} & \multicolumn{2}{c|}{\bf{SUN}} & \multicolumn{2}{c|}{\bf{Places}} & \multicolumn{2}{c|}{\bf{Textures}} & \multicolumn{2}{c|}{\bf{OpenImage-O}} & \multicolumn{2}{c|}{\bf{ImageNet-O}}   &  \multicolumn{2}{c}{\cellcolor[gray]{0.9}\bf{Average}}  \\ 
\hhline{~*{12}{-}|*{2}{-}}
& AU$\uparrow$          & FP$\downarrow$         & AU$\uparrow$         & FP$\downarrow$   & AU$\uparrow$         & FP$\downarrow$      & AU$\uparrow$           & FP$\downarrow$ & AU$\uparrow$          & FP$\downarrow$
                          & AU$\uparrow$           & FP$\downarrow$ & AU$\uparrow$           & FP$\downarrow$ \\ \hline 
\textit{MSP} & 89.04 & 49.30 & 82.99 & 61.46 & 81.95 & 64.91 & 84.33 & 57.32 & 86.25 & 55.89 & 57.62 & 92.10 & 80.36 & 63.50 \\
\textit{MaxLogit} & 90.54 & 37.76 & 80.88 & 55.30 & 78.52 & 61.36 & 84.81 & 45.18 & 82.97 & 51.04 & 45.37 & 93.05 & 77.18 & 57.28 \\
\textit{ODIN} & 75.26 & 75.28 & 60.07 & 83.23 & 55.45 & 86.82 & 75.86 & 68.35 & 64.37 & 82.93 & 56.56 & 92.15 & 64.60 & 81.46 \\
\textit{Energy} & 88.79 & 43.36 & 76.62 & 62.14 & 73.14 & 68.26 & 82.78 & 46.37 & 76.69 & 59.99 & 40.03 & 94.65 & 73.01 & 62.46 \\
\textit{GEN} & 94.41 & 28.58 & 87.36 & \textbf{50.14} & 85.44 & \textbf{56.12} & 89.07 & 42.70 & 91.18 & 40.67 & 65.53 & \underline{89.10} & 85.50 & 51.22 \\
\textit{ReAct} & 80.15 & 68.51 & 62.81 & 81.64 & 58.01 & 86.14 & 74.11 & 63.78 & 66.68 & 76.46 & 40.92 & 96.90 & 63.78 & 78.91 \\
\textit{DICE} & 80.47 & 66.61 & 67.78 & 76.60 & 63.51 & 83.42 & 75.38 & 60.53 & 63.90 & 75.45 & 28.18 & 98.10 & 63.20 & 76.78 \\
\textit{ViM} & \underline{95.26} & 27.62 & 80.81 & 63.22 & 75.83 & 67.38 & \underline{91.30} & 41.21 & 93.06 & 37.38 & \textbf{75.79} & \textbf{84.55} & 85.34 & 53.56 \\
\textit{ASH-S} & 47.01 & 90.66 & 65.40 & 80.89 & 65.26 & 84.48 & 65.48 & 73.95 & 49.06 & 89.75 & 39.32 & 98.05 & 55.26 & 86.30 \\
\textit{OptFS} & 81.08 & 63.02 & 79.33 & 68.12 & 78.40 & 71.79 & 77.54 & 66.22 & 79.82 & 65.97 & 54.35 & 89.95 & 75.09 & 70.84 \\
\hline
\textit{CARef} & 95.11 & \underline{26.30} & \underline{87.61} & 54.21 & \underline{85.49} & 59.71 & 91.18 & \underline{38.62} & \textbf{93.30} & \underline{35.97} & \underline{73.52} & 89.70 & \underline{87.70} & \underline{50.75} \\
\textit{CADRef} & \textbf{95.35} & \textbf{24.23} & \textbf{87.82} & \underline{52.19} & \textbf{85.68} & \underline{58.26} & \textbf{91.38} & \textbf{37.06} & \underline{93.29} & \textbf{35.75} & 72.87 & 89.45 & \textbf{87.73} & \textbf{49.49} \\
\Xhline{1px}
\end{NiceTabular}
\caption{Results of OOD detection on ImageNet-1k benchmark with MaxViT-T. 
$\uparrow$ indicates that higher values are better, while $\downarrow$ indicates that lower values are better.
All values are percentages, with the best and second-best results being \textbf{highlighted} and \underline{underlined}, respectively.}
\label{maxvit_result}
\end{table*}

\end{document}